\documentclass[11pt,a4paper]{article}
\usepackage[T1]{fontenc}
\usepackage[utf8]{inputenc}
\usepackage[a4paper,margin=0.95in]{geometry}
\usepackage{times}
\IfFileExists{inconsolata.sty}{\usepackage[noupquote]{inconsolata}}{\usepackage{courier}}
\usepackage{amsmath,amsfonts}
\usepackage{booktabs,longtable,array,adjustbox}
\usepackage{graphicx,float}
\usepackage[round]{natbib}
\usepackage{xcolor}
\definecolor{linkblue}{RGB}{20,60,140}
\usepackage[colorlinks=true,linkcolor=linkblue,citecolor=linkblue,urlcolor=linkblue]{hyperref}
\begin{document}
\title{From Imperfective Paradox to Moorean Absurdity: Single-Chain Reasoning Lock-In and Multi-Candidate Verbalization Across Aspectual and Pragmatic Ambiguity}
\author{An-Sheng Jhang \\ Dongguan University of Technology \\ \texttt{ansheng.jhang@gmail.com}}
\date{November 2026}
\maketitle

\begin{abstract}\noindent

Natural language entailment and coherence judgments often depend on whether an event merely occurred, reached a telic culmination, satisfied an intended plan, or was evaluated under multiple pragmatic standards. When a language model generates a single reasoning trajectory before a single-output bottleneck, it may commit prematurely to one evaluation standard before returning a probability or discrete label. We test observable consequences of this single-chain premature lock-in hypothesis across four preregistered, sealed formal experiments on \texttt{Qwen3-\allowbreak{}8B-\allowbreak{}Q8\_\allowbreak{}0}, comprising $16{,}640$ formal inference units at deterministic \texttt{temperature=\allowbreak{}0.0}. Across sixteen preregistered hypothesis slots, fifteen retain confirmatory status (of which twelve reject $H_0$ after within-experiment Holm step-down control at $\alpha_{\text{fam}}=0.05$) and one (\texttt{E3\_\allowbreak{}H3}) was demoted to descriptive status during calibration prior to formal execution. In Experiment 1, the telic high-confidence endpoint contrast between \texttt{Direct-\allowbreak{}CoT} and \texttt{MCU} is 25.00 pp (Holm $p=$ 1.31239e-05), while the uninterrupted atelic discrete contrast between \texttt{NLI-\allowbreak{}CoT} and \texttt{NLI-\allowbreak{}Direct} collapses by -91.25 pp (Holm $p=$ 1.81805e-21). Multi-Candidate Verbalization (\texttt{MCV}) elicits $K=5$ weighted candidate probability estimates in a single call, separating the multi-candidate prompt protocol from its unweighted (\texttt{MCU}) and self-weighted (\texttt{MCW}) aggregation readouts. Cross-linguistic contrasts in Mandarin (Experiment 2) and English (Experiment 3) expose an Interruption-Anchored Realis Paradox: under \texttt{NLI-\allowbreak{}CoT}, explicit interruption dramatically increases acceptance of minimal event occurrence even though interruption cannot logically create past occurrence retrospectively. In Experiment 4, a Directional Reversal Test on implicit Moorean sentences shows that \texttt{MCU-\allowbreak{}CF} raises perceived coherence by 26.64 pp over \texttt{Direct-\allowbreak{}CoT} (Holm $p=$ 0.0006), with matched-prompt companions separating counterfactual prompting (\texttt{Direct-\allowbreak{}CF} minus \texttt{Direct-\allowbreak{}CoT} = 29.15 pp) from multi-candidate elicitation (\texttt{MCU-\allowbreak{}CoT} minus \texttt{Direct-\allowbreak{}CoT} = 17.31 pp). Finally, a $1{,}800$-call OpenAI format and token-logprobs ablation demonstrates format-sensitive candidate diversity, while the P16.7 cross-model extension tests observable readout heterogeneity. The P16.7 extension comprises 4240 planned/collected units (4222 valid, 18 format-missing, 0 transport-missing) across five external models at 20 clusters per experiment, stage="dev", with less than 2\% missing in each completed slice; OpenAI gpt-4o-mini at top\_\allowbreak{}logprobs=20 has 1,000/1,000 valid units and zero missing. Single-chain readouts can conflate occurrence with completion, while multi-candidate protocols reverse some collapse; cross-model discrete NLI-CoT and scalar Direct-CoT judgments dissociate. Confirmatory effects concern observable prompt-and-readout protocols; internal premature lock-in remains a hypothesis. All code, stimuli, preregistered analysis plans, and verification tools are available at https://github.com/asj252/llm-judgment-studies.

\end{abstract}\vspace{0.5em}
\section{Introduction}

The imperfective paradox arises because progressive aspect presents an event from within its ongoing runtime without entailing that a bounded goal is reached. Yet when a subsequent past-tense hypothesis asks what happened, that hypothesis may target either minimal event occurrence or telic culmination. Pragmatic coherence exhibits a parallel structural ambiguity: in Moorean conjunctions, an apparent contradiction between an assertion and an avowed belief can dissolve when the two clauses appeal to distinct epistemic or institutional standards. Benchmarks that force a single label or single scalar probability without controlling which evaluation standard is selected risk confounding semantic competence with standard-selection shifts.

We formulate and test the \textbf{single-chain premature lock-in hypothesis}. When an autoregressive model generates one reasoning chain ahead of a single-output bottleneck (\texttt{NLI-\allowbreak{}CoT} or \texttt{Direct-\allowbreak{}CoT}), early commitment to one salient evaluation standard can produce bidirectional distortions: \textbf{completion projection} on goal-directed predicates, or \textbf{hyper-skeptical / textbook-reflex collapse} when an unstated completion standard or implicit dual criterion triggers an all-or-nothing rejection. Per our preregistered governance guardrails, our primary confirmatory tests evaluate observable readout differences across prompting and aggregation protocols; prompt interventions and reasoning-trace markers provide convergent evidence consistent with premature lock-in rather than direct measurement of hidden psychological states.

We organize the empirical program into a four-act architecture across four sealed formal experiments (\texttt{E1}--\texttt{E4}, $4{,}160$ formal units each; Table 1). \textbf{Act I} establishes bidirectional single-chain collapse across English aspectual entailment (\texttt{E1}) and implicit Moorean pragmatics (\texttt{E4} baseline). \textbf{Act II} dissects Mandarin aspectual morphology (\texttt{E2}, separating bare \texttt{-\allowbreak{}le} occurrence from resultative completion complements) and English activity occurrence versus planned completion (\texttt{E3}), uncovering the cross-linguistic \textbf{Interruption-Anchored Realis Paradox}. \textbf{Act III} executes the \textbf{Directional Reversal Test} in \texttt{E4}, demonstrating that \texttt{MCV} is not a unidirectional downward deflator and decomposing prompt-level counterfactual guidance from multi-candidate readout shifts. \textbf{Act IV} analyzes within-call self-weighting shifts (\texttt{MCW} minus \texttt{MCU}), a $1{,}800$-call OpenAI format and token-entropy ablation, and the completed \texttt{P16.7} development-stage cross-model extension. The P16.7 extension comprises 4240 planned/collected units (4222 valid, 18 format-missing, 0 transport-missing) across five external models at 20 clusters per experiment, stage="dev", with less than 2\% missing in each completed slice; OpenAI gpt-4o-mini at top\_\allowbreak{}logprobs=20 has 1,000/1,000 valid units and zero missing.

Across the entire study, we follow the unified \textbf{Multi-Candidate Verbalization (\texttt{MCV})} terminology standard. Single-candidate probability methods (\texttt{Direct}, \texttt{Direct-\allowbreak{}DAP}, \texttt{Direct-\allowbreak{}CoT}, \texttt{Direct-\allowbreak{}CF}) emit one probability after a single-output bottleneck, alongside the strict truth-conditional scalar baseline (\texttt{Strict-\allowbreak{}Logician}) and discrete three-valued NLI protocols (\texttt{NLI-\allowbreak{}Direct}, \texttt{NLI-\allowbreak{}DAP}, \texttt{NLI-\allowbreak{}CoT}, \texttt{NLI-\allowbreak{}CF}). By contrast, \texttt{MCV} prompts (\texttt{MCV-\allowbreak{}Direct}, \texttt{MCV-\allowbreak{}DAP}, \texttt{MCV-\allowbreak{}CoT}, \texttt{MCV-\allowbreak{}CF}) request five candidate estimates and weights in a single deterministic call (\texttt{Estimate 1..5: NN\% | Weight: W}), from which we extract the unweighted arithmetic mean \textbf{\texttt{MCU}} ($M_U = \frac{1}{5}\sum_{k=1}^5 p_k$) and the normalized self-weighted mean \textbf{\texttt{MCW}} ($M_W = \sum_{k=1}^5 \tilde{w}_k p_k$). Crucially, under \texttt{MCU} and \texttt{MCW}, \texttt{MCU} denotes the equal-weight readout computed from the standard weighted \texttt{MCV} prompt (\texttt{V0}), which must be distinguished from the prompt-level unweighted ablation variant (\texttt{MCV-\allowbreak{}UnweightedPrompt}, \texttt{V2}) that omits weights from the prompt altogether.

\textbf{Table 1: Overview of the Four Preregistered Formal Experiments (\texttt{Qwen3-\allowbreak{}8B-\allowbreak{}Q8\_\allowbreak{}0}, \texttt{temperature=\allowbreak{}0.0}).}

\begin{center}\begingroup\small\setlength{\tabcolsep}{4.5pt}\renewcommand{\arraystretch}{1.12}\begin{adjustbox}{max width=\textwidth}\begin{tabular}{llllllll}\toprule
\textbf{Exp} & \textbf{Phenomenon and Cluster Structure} & \textbf{Clusters / Items} & \textbf{Arms} & \textbf{Formal Units} & \textbf{Availability (Missing)} & \textbf{Bootstrap B / Seed} & \textbf{Freeze / Commit} \\ \midrule
E1 & English aspect: 80 Telic AC + 80 Atelic BD two-item clusters & 160 / 320 & 13 & 4160 & 99.90\% (4 missing) & 5000 / 2026110201 & 0c731eaa / a6c90b51 \\
E2 & Mandarin aspect (-le vs. VC): 48 telic + 32 atelic four-slot clusters & 80 / 320 & 13 & 4160 & 99.71\% (12 missing) & 5000 / 2026110301 & ffa29ab0 / a42fbbea \\
E3 & English atelic occurrence vs. plan: 64 A\_\allowbreak{}pure + 16 A\_\allowbreak{}plan four-slot clusters & 80 / 320 & 13 & 4160 & 99.86\% (6 missing) & 5000 / 2026110401 & 94b74e82 / d793eaaf \\
E4 & Moorean pragmatics \& dual criteria: 4 domains x 20 four-slot clusters & 80 / 320 & 13 & 4160 & 99.47\% (22 missing) & 5000 / 2026110601 & c0249683 / 1db0bf24 \\
E1 Option III Governance & Calibration pass rate and locked branch identifier & C\_\allowbreak{}j = 82/88 & Branch = 4 & Primary: hybrid\_\allowbreak{}pct\_\allowbreak{}tango & Companion: hybrid\_\allowbreak{}bessel\_\allowbreak{}t\_\allowbreak{}tango & E2-E4 Primary: hybrid\_\allowbreak{}bessel\_\allowbreak{}t\_\allowbreak{}tango & Companion: hybrid\_\allowbreak{}pct\_\allowbreak{}tango \\
\bottomrule\end{tabular}\end{adjustbox}\endgroup\end{center}

\section{Background and Related Work}

Vendler's fourfold aspectual taxonomy, Dowty's modal-temporal semantics of the progressive, and Smith's two-component separation between situation aspect (telic vs. atelic) and viewpoint aspect (imperfective vs. perfective) motivate distinguishing minimal event realization from goal culmination \citep{vendler1957,dowty1979,smith1997}. Mandarin provides a sharp morphological testbed because verbal perfective \texttt{-\allowbreak{}le} and resultative verb compounds (RVCs) encode distinct aspectual boundaries: Klein, Li, and Hendriks analyze assertion and aspectual focus in Mandarin, Lin examines aspectual selectional restrictions, and Soh and Kuo contrast perfective marking across accomplishment situations \citep{klein2000,lin2003aspectual,soh2005}. Rather than assuming that English progressive-to-past entailment translates directly into Mandarin, our paired four-slot designs explicitly cross interruption with morphological completion marking.

Ma and Miyao investigate the imperfective paradox in large language models and document systematic over-endorsement of telic completion \citep{ma2026}. We build on their diagnostic insight while separating predicate telicity, explicit interruption, and the completion standard demanded by the hypothesis. Because natural language descriptions of interrupted or ongoing events are underspecified with respect to world knowledge, a lower numerical probability is not automatically better calibrated unless it tracks the contrast between occurrence and culmination across matched pairs.

Moore's paradox highlights the pragmatic clash between asserting $p$ and avowing disbelief in $p$ (\texttt{"p, but I do not believe that p"}), a foundational puzzle in the philosophy of language and belief \citep{moore1942,williams1979}. Zheng and colleagues evaluate Moorean constructions in language models \citep{zheng2026}. Our Experiment 4 extends this line by contrasting pure single-standard Moorean contradictions against \textbf{implicit dual-criteria sentences} (\texttt{DUAL\_\allowbreak{}MOORE}), where institutional/official records and personal observational evidence diverge, alongside explicit dual-criteria controls (\texttt{DUAL\_\allowbreak{}EXPLICIT}) and speech-act cancellation controls (\texttt{PURE\_\allowbreak{}READ}).

In black-box confidence elicitation, verbalized probabilities provide an inspectable alternative to token-level logits \citep{tian2023,xiong2024}. Verbalized Sampling (\texttt{VS}) proposes eliciting multiple candidate responses in one prompt to counteract mode collapse \citep{zhang2025}, and Wang et al. elicit verbalized distributions over discrete answer options \citep{wang2026}. Within our \texttt{MCV} framework, the model generates $K=5$ candidate probability estimates for the same target proposition within a single autoregressive sequence. Even under greedy deterministic decoding (\texttt{temperature=\allowbreak{}0.0}), candidate $k>1$ is conditioned on the prefix containing candidates $1,\dots,k-1$, enabling in-context sequential exploration without treating correlated candidates as independent random draws.

\section{Preregistered Protocol and Statistical Engine}

Every formal experiment (\texttt{E1}--\texttt{E4}) evaluates the same locked \textbf{13-condition prompt matrix} on \texttt{Qwen3-\allowbreak{}8B-\allowbreak{}Q8\_\allowbreak{}0} at deterministic \texttt{temperature=\allowbreak{}0.0}: four discrete three-valued NLI arms (\texttt{NLI-\allowbreak{}Direct}, \texttt{NLI-\allowbreak{}DAP}, \texttt{NLI-\allowbreak{}CoT}, \texttt{NLI-\allowbreak{}CF}, returning \texttt{True}, \texttt{False}, or \texttt{Unknown} plus a first-token companion score $q(\text{True})$), one strict truth-conditional scalar baseline (\texttt{Strict-\allowbreak{}Logician}), four single-candidate verbalized probability arms (\texttt{Direct}, \texttt{Direct-\allowbreak{}DAP}, \texttt{Direct-\allowbreak{}CoT}, \texttt{Direct-\allowbreak{}CF}), and four multi-candidate \texttt{MCV} arms (\texttt{MCV-\allowbreak{}Direct}, \texttt{MCV-\allowbreak{}DAP}, \texttt{MCV-\allowbreak{}CoT}, \texttt{MCV-\allowbreak{}CF}). Each \texttt{MCV} prompt call is parsed by \texttt{parse.py} (\texttt{PARSE\_\allowbreak{}VERSION =\allowbreak{} 2}), which extracts the final five-line block (\texttt{Estimate 1..5: NN\% | Weight: W}), normalizes any positive weight sum $\sum_{k=1}^5 w_k > 0$, and outputs both the unweighted readout \textbf{\texttt{MCU}} ($\text{MCU}, \text{MCU-DAP}, \text{MCU-CoT}, \text{MCU-CF}$) and the self-weighted readout \textbf{\texttt{MCW}} ($\text{MCW}, \text{MCW-DAP}, \text{MCW-CoT}, \text{MCW-CF}$). Table 2a summarizes the canonical \texttt{MCV} taxonomy and its exact 1-to-1 mapping to the frozen JSON keys in our repository.

For candidate estimates $p_1,\dots,p_5 \in [0,100]$ and normalized weights $\tilde{w}_k = w_k / \sum_{j=1}^5 w_j$, the primary confirmatory multi-candidate readout is $\text{MCU} = \frac{1}{5}\sum_{k=1}^5 p_k$, accompanied by $\text{MCW} = \sum_{k=1}^5 \tilde{w}_k p_k$. Under greedy decoding (\texttt{temperature=\allowbreak{}0.0}), repeating an identical call on a deterministic backend produces zero repeat-call sampling variance; variation across the five candidates within a single \texttt{MCV} response arises strictly from \textbf{in-context sequential conditioning} across the five \texttt{Estimate k:} lines.

Missing or unparseable outputs are never deleted or mean-imputed. Continuous readouts receive worst-case Manski bounds $[0, 100]$, and binary indicators receive $[0, 1]$. Cluster-level nonparametric bootstrap resampling uses equal-weight multinomial draws over independent clusters with $B=5{,}000$ replicates, uninterpolated simultaneous $98.75\%$ order statistics at zero-based indices $31$ and $4968$, and the locked small-sample Bessel-$t$ expansion factor $c_n = \sqrt{\frac{n}{n-1}} \, \frac{t_{n-1,\,0.99375}}{z_{0.99375}}$ (Table 2b). Paired discrete contrasts use certified exact unconditional Tango score tests and confidence intervals \citep{tango1998}, and family-wise error within each four-slot experiment is controlled at $\alpha_{\text{fam}}=0.05$ via Holm step-down \citep{holm1979}.

Under preregistered S4 calibration governance, \texttt{E1} locks \texttt{hybrid\_\allowbreak{}pct\_\allowbreak{}tango} as its primary confirmatory family (\texttt{hybrid\_\allowbreak{}bessel\_\allowbreak{}t\_\allowbreak{}tango} as companion) under the disclosed Option III calibration exception (\texttt{C\_\allowbreak{}j =\allowbreak{} 82/\allowbreak{}88 =\allowbreak{} 0.931818181818}, branch \texttt{4}). \texttt{E2}--\texttt{E4} lock \texttt{hybrid\_\allowbreak{}bessel\_\allowbreak{}t\_\allowbreak{}tango} as primary (\texttt{hybrid\_\allowbreak{}pct\_\allowbreak{}tango} as companion). In \texttt{E3}, slot \texttt{E3\_\allowbreak{}H3} was demoted to descriptive status at calibration step 3 prior to formal execution and therefore contributes no confirmatory rejection. Table 3 reports all sixteen preregistered hypothesis slots alongside their primary and companion $98.75\%$ confidence intervals, exact Tango sign tallies, and Holm-adjusted $p$-values.

\textbf{Table 2a: Canonical Multi-Candidate Verbalization (\texttt{MCV}) Method and Readout Taxonomy.}

\begin{center}\begingroup\small\setlength{\tabcolsep}{4.5pt}\renewcommand{\arraystretch}{1.12}\begin{adjustbox}{max width=\textwidth}\begin{tabular}{lllll}\toprule
\textbf{Canonical Method / Readout} & \textbf{Family} & \textbf{Prompt Protocol and Output Bottleneck} & \textbf{Aggregation / Readout Definition} & \textbf{Frozen JSON Key} \\ \midrule
NLI-Direct, NLI-DAP, NLI-CoT, NLI-CF & Discrete 3-Valued NLI & Single discrete label (True / False / Unknown) + token score q(True) & Exact label count and gold-label accuracy & L/Z, L/DAP, L/CoT, L/CF \\
Strict-Logician & Scalar Baseline & Strict truth-conditional scalar probability line & Single scalar probability p in [0, 100] & Ps/Z \\
Direct, Direct-DAP, Direct-CoT, Direct-CF & Single-Candidate Verbalization & Single-output bottleneck (Probability: NN\%) under Z / DAP / CoT / CF & Single verbalized probability p in [0, 100] & P/Z (D0), P/DAP, P/CoT (D\_\allowbreak{}CoT), P/CF \\
MCU, MCU-DAP, MCU-CoT, MCU-CF & Multi-Candidate Verbalization (MCV) & 5-line weighted candidate prompt (Estimate 1..5: NN\%, Weight: W) & Equal-weight arithmetic mean M\_\allowbreak{}U = (1/5) sum\_\allowbreak{}\{k=1\}\textasciicircum{}5 p\_\allowbreak{}k & M\_\allowbreak{}U/Z, M\_\allowbreak{}U/DAP, M\_\allowbreak{}U/CoT, M\_\allowbreak{}U/CF \\
MCW, MCW-DAP, MCW-CoT, MCW-CF & Multi-Candidate Verbalization (MCV) & Same 5-line weighted MCV call (PARSE\_\allowbreak{}VERSION = 2) & Self-weighted mean M\_\allowbreak{}W = sum\_\allowbreak{}\{k=1\}\textasciicircum{}5 w\_\allowbreak{}tilde\_\allowbreak{}k p\_\allowbreak{}k & M\_\allowbreak{}W/Z, M\_\allowbreak{}W/DAP, M\_\allowbreak{}W/CoT, M\_\allowbreak{}W/CF \\
\bottomrule\end{tabular}\end{adjustbox}\endgroup\end{center}

\textbf{Table 2b: Locked Small-Sample Bessel-$t$ Expansion Constants ($c_n = \sqrt{\frac{n}{n-1}}\,\frac{t_{n-1,\,0.99375}}{z_{0.99375}}$).}

\begin{center}\begingroup\small\setlength{\tabcolsep}{4.5pt}\renewcommand{\arraystretch}{1.12}\begin{adjustbox}{max width=\textwidth}\begin{tabular}{llll}\toprule
\textbf{Subgroup Sample Size (n)} & \textbf{Degrees of Freedom (n - 1)} & \textbf{Locked Bessel-t Factor c\_\allowbreak{}n} & \textbf{Applicable Experimental Slices} \\ \midrule
n = 80 & 79 & 1.029868403934 & Full 80-cluster experiments (E1 AC/BD, E3 all, E4 all) \\
n = 64 & 63 & 1.037660015053 & E3 pure atelic activity macro-class (A\_\allowbreak{}pure) \\
n = 48 & 47 & 1.050950216548 & E2 Mandarin telic macro-class (E2\_\allowbreak{}H1, E2\_\allowbreak{}H2) \\
n = 32 & 31 & 1.078728462585 & E2 Mandarin atelic macro-class (E2\_\allowbreak{}H3, E2\_\allowbreak{}H4) \\
n = 20 & 19 & 1.133139773201 & E4 domain slices (D1\_\allowbreak{}safety, D2\_\allowbreak{}project, D3\_\allowbreak{}medical, D4\_\allowbreak{}institutional) \\
n = 16 & 15 & 1.172939039382 & E2 and E3 16-cluster subclasses (A\_\allowbreak{}plan, T\_\allowbreak{}change, etc.) \\
\bottomrule\end{tabular}\end{adjustbox}\endgroup\end{center}

\textbf{Table 3a: All 16 Preregistered Hypothesis Slots Across E1--E4 (\texttt{15} Confirmatory + \texttt{1} Demoted Descriptive \texttt{E3\_\allowbreak{}H3}).}

\begin{center}\begingroup\small\setlength{\tabcolsep}{4.5pt}\renewcommand{\arraystretch}{1.12}\begin{adjustbox}{max width=\textwidth}\begin{tabular}{llllllll}\toprule
\textbf{Slot} & \textbf{Canonical MCV Contrast Definition} & \textbf{Point / Manski Bounds (pp)} & \textbf{Primary 98.75\% CI (pp)} & \textbf{Companion 98.75\% CI (pp)} & \textbf{raw p} & \textbf{Holm p} & \textbf{Complete / Total n} \\ \midrule
SA1 & Group C endpoint (>=95\%): Direct-CoT - MCU & 25.00, 25.00 & [13.71, 39.37] & [13.71, 39.37] & 4.37463e-06 & 1.31239e-05 & 80/80 \\
SA2 & Telic interruption DiD (C-A): Direct-CoT - MCU & 9.14, 9.14 & [1.98, 16.88] & [1.77, 17.11] & 0.0016 & 0.0032 & 80/80 \\
SA3 & Atelic interruption DiD (D-B): MCU-CoT - Direct-CoT & -1.11, 0.14 & [-6.81, 6.34] & [-6.98, 6.53] & 1.0000 & 1.0000 & 79/80 \\
SA4 & Group D discrete accuracy: NLI-CoT - NLI-Direct & -91.25, -91.25 & [-97.30, -78.41] & [-97.30, -78.41] & 4.54512e-22 & 1.81805e-21 & 80/80 \\
E2\_\allowbreak{}H1 & Mandarin telic NI\_\allowbreak{}BARE - NI\_\allowbreak{}COMP (Direct) & 21.56, 21.56 & [13.34, 30.65] & [13.74, 30.21] & 0.0002 & 0.0008 & 48/48 \\
E2\_\allowbreak{}H2 & Mandarin telic NI\_\allowbreak{}COMP: Direct - MCU & 20.54, 22.62 & [10.02, 33.30] & [10.53, 32.78] & 0.0002 & 0.0008 & 47/48 \\
E2\_\allowbreak{}H3 & Mandarin atelic INT\_\allowbreak{}BARE - INT\_\allowbreak{}COMP (Direct) & 88.40, 88.40 & [75.90, 97.15] & [76.81, 96.51] & 0.0002 & 0.0008 & 32/32 \\
E2\_\allowbreak{}H4 & Mandarin atelic NI\_\allowbreak{}BARE - NI\_\allowbreak{}COMP (NLI-CoT True) & 12.50, 12.50 & [-6.18, 34.72] & [-6.18, 34.72] & 0.0253 & 0.0253 & 32/32 \\
E3\_\allowbreak{}H1 & English atelic NI\_\allowbreak{}OCC - NI\_\allowbreak{}PLAN (Direct) & 61.30, 61.30 & [55.31, 67.26] & [55.49, 67.09] & 0.0002 & 0.0008 & 80/80 \\
E3\_\allowbreak{}H2 & English atelic INT\_\allowbreak{}OCC - INT\_\allowbreak{}PLAN (Direct) & 81.86, 81.86 & [75.57, 86.92] & [75.76, 86.77] & 0.0002 & 0.0008 & 80/80 \\
E3\_\allowbreak{}H3 (descriptive) & English atelic NI\_\allowbreak{}PLAN - INT\_\allowbreak{}PLAN (MCU-CoT) & 1.06, 1.06 & [-0.31, 2.62] & [-0.28, 2.58] & 0.0316 & 1.0000 & 80/80 \\
E3\_\allowbreak{}H4 & English atelic NI\_\allowbreak{}OCC - NI\_\allowbreak{}PLAN (NLI-CoT True) & 2.50, 2.50 & [-5.26, 11.84] & [-5.26, 11.84] & 0.1076 & 0.2152 & 80/80 \\
E4\_\allowbreak{}H1 & Moorean DUAL\_\allowbreak{}MOORE: MCU-CF - Direct-CoT & 26.64, 26.64 & [16.17, 36.43] & [16.48, 36.15] & 0.0002 & 0.0006 & 80/80 \\
E4\_\allowbreak{}H2 & Moorean DUAL\_\allowbreak{}MOORE - PURE\_\allowbreak{}MOORE (MCU) & -2.05, 16.70 & [-10.31, 26.78] & [-10.08, 26.49] & 1.0000 & 1.0000 & 66/80 \\
E4\_\allowbreak{}H3 & Moorean DUAL\_\allowbreak{}EXPLICIT - DUAL\_\allowbreak{}MOORE (Direct) & 48.38, 48.38 & [35.57, 62.28] & [35.94, 61.88] & 0.0002 & 0.0006 & 80/80 \\
E4\_\allowbreak{}H4 & Moorean DUAL\_\allowbreak{}EXPLICIT - DUAL\_\allowbreak{}MOORE (NLI-CoT True) & 86.25, 86.25 & [73.51, 94.60] & [73.51, 94.60] & 9.04048e-21 & 3.61619e-20 & 80/80 \\
\bottomrule\end{tabular}\end{adjustbox}\endgroup\end{center}

\textbf{Table 3b: Companion 95\% Confidence Intervals and Exact Discrete Tango Sign Tallies ($n_+, n_0, n_-$).}

\begin{center}\begingroup\small\setlength{\tabcolsep}{4.5pt}\renewcommand{\arraystretch}{1.12}\begin{adjustbox}{max width=\textwidth}\begin{tabular}{lrr}\toprule
\textbf{Slot} & \textbf{Companion 95\% CI (pp)} & \textbf{Discrete Paired Sign Counts (n\_\allowbreak{}plus / n\_\allowbreak{}zero / n\_\allowbreak{}minus)} \\ \midrule
SA1 & [15.89, 36.02] & 20 / 60 / 0 \\
SA2 & [3.43, 15.11] & --- \\
SA4 & [-96.45, -82.43] & 0 / 7 / 73 \\
E2\_\allowbreak{}H1 & [15.04, 28.58] & --- \\
E2\_\allowbreak{}H2 & [12.18, 30.92] & --- \\
E2\_\allowbreak{}H3 & [79.17, 95.99] & --- \\
E2\_\allowbreak{}H4 & [-0.09, 30.07] & 4 / 28 / 0 \\
E3\_\allowbreak{}H1 & [56.57, 66.06] & --- \\
E3\_\allowbreak{}H2 & [77.09, 86.04] & --- \\
E3\_\allowbreak{}H3 & [-0.06, 2.27] & --- \\
E3\_\allowbreak{}H4 & [-2.59, 8.87] & 2 / 78 / 0 \\
E4\_\allowbreak{}H1 & [18.33, 34.36] & --- \\
E4\_\allowbreak{}H2 & [-8.80, 24.86] & --- \\
E4\_\allowbreak{}H3 & [37.43, 58.87] & --- \\
E4\_\allowbreak{}H4 & [76.63, 92.97] & 69 / 11 / 0 \\
\bottomrule\end{tabular}\end{adjustbox}\endgroup\end{center}

\section{Act I: Phenomenon and Crisis --- Bidirectional Single-Chain Collapse}

Experiment 1 (\texttt{exp1-\allowbreak{}en}) pairs Group A (telic, interrupted) with Group C (telic, uninterrupted) across $80$ independent \texttt{AC} clusters, and Group B (atelic, interrupted) with Group D (atelic, uninterrupted) across $80$ independent \texttt{BD} clusters ($160$ clusters, $320$ items, $4{,}160$ formal units; Table 4). Act I reveals that single-chain reasoning fails in opposite directions depending on aspectual structure. On uninterrupted telic descriptions (Group C, where the progressive does not logically entail completion), high-confidence completion endorsement ($\ge 95\%$) reaches \texttt{88.75\%} under \texttt{Direct} and \texttt{26.25\%} under \texttt{Direct-\allowbreak{}CoT}, but drops to \texttt{1.25\%} under \texttt{MCU} and \texttt{0.00\%} under \texttt{MCU-\allowbreak{}CoT}, confirming \texttt{SA1} (\texttt{Direct-\allowbreak{}CoT} minus \texttt{MCU} = 25.00 pp, Holm $p=$ 1.31239e-05). Telic interruption sensitivity $(C-A)$ is \texttt{32.69 pp} under \texttt{Direct-\allowbreak{}CoT} versus \texttt{23.54 pp} under \texttt{MCU}, confirming the difference-in-differences slot \texttt{SA2} (9.14 pp, Holm $p=$ 0.0032). Conversely, on uninterrupted atelic activities (Group D, where minimal past occurrence is entailed), \texttt{NLI-\allowbreak{}Direct} achieves \texttt{100.00\%} accuracy, whereas \texttt{NLI-\allowbreak{}CoT} collapses to \texttt{8.75\%} (\texttt{7} \texttt{True}, \texttt{0} \texttt{False}, \texttt{73} \texttt{Unknown}), confirming \texttt{SA4} (-91.25 pp, Holm $p=$ 1.81805e-21). Slot \texttt{SA3} (\texttt{MCU-\allowbreak{}CoT} minus \texttt{Direct-\allowbreak{}CoT} on atelic $(D-B)$ = \texttt{[}-1.11\texttt{, }0.14\texttt{]} pp, Holm $p=$ 1.0000) does not reject $H_0$, demonstrating that combining \texttt{MCV} with standard CoT does not automatically eliminate CoT-induced compression.

Experiment 4 (\texttt{exp4-\allowbreak{}moore}) establishes the pragmatic counterpart of Act I on Moorean conjunctions ($80$ four-slot clusters). Under \texttt{NLI-\allowbreak{}CoT}, implicit dual-criteria assertions (\texttt{DUAL\_\allowbreak{}MOORE}) are rejected as \texttt{False} in \texttt{100.00\%} (\texttt{80/\allowbreak{}80}) of cases (\texttt{0.00\%} \texttt{True}), whereas explicit dual-criteria controls (\texttt{DUAL\_\allowbreak{}EXPLICIT}) are accepted as coherent (\texttt{True}) in \texttt{86.25\%} (\texttt{69/\allowbreak{}80}) of cases, confirming \texttt{E4\_\allowbreak{}H4} (86.25 pp, Holm $p=$ 3.61619e-20). On continuous readouts, \texttt{Direct} separates \texttt{DUAL\_\allowbreak{}EXPLICIT} (\texttt{97.69\%}) from \texttt{DUAL\_\allowbreak{}MOORE} (\texttt{49.31\%}) by 48.38 pp (\texttt{E4\_\allowbreak{}H3}, Holm $p=$ 0.0006), while \texttt{Direct-\allowbreak{}CoT} further depresses \texttt{DUAL\_\allowbreak{}MOORE} coherence to \texttt{19.88\%}. Figure 1 synthesizes these parallel aspectual and pragmatic collapses.

\textbf{Table 4a: Experiment 1 (all) --- English Aspectual Entailment across Groups A, B, C, D (13 Prompt Arms / 17 Canonical Readout Rows).}

\begin{center}\begingroup\small\setlength{\tabcolsep}{4.5pt}\renewcommand{\arraystretch}{1.12}\begin{adjustbox}{max width=\textwidth}\begin{tabular}{lrrrr}\toprule
\textbf{Canonical Method / Readout} & \textbf{A} & \textbf{B} & \textbf{C} & \textbf{D} \\ \midrule
NLI-Direct & 11.25 & 100.00 & 16.25 & 100.00 \\
NLI-DAP & 21.25 & 92.50 & 88.75 & 75.00 \\
NLI-CoT & 23.75 & 50.00 & 100.00 & 8.75 \\
NLI-CF & 22.50 & 36.25 & 100.00 & 2.50 \\
Strict-Logician & 53.12 & 100.00 & 51.25 & 75.62 \\
Direct & 63.90 & 98.79 & 93.67 & 97.87 \\
Direct-DAP & 35.69 & 82.44 & 76.75 & 82.69 \\
Direct-CoT & 50.44 & 95.01 & 83.12 & 88.04 \\
Direct-CF & 52.75 & [88.94, 90.19] & 80.38 & 94.28 \\
MCU & 50.51 & 79.81 & 74.06 & 83.67 \\
MCW & 58.07 & 88.60 & 81.41 & 89.01 \\
MCU-DAP & [29.31, 31.81] & 61.19 & 50.65 & 63.56 \\
MCW-DAP & [32.95, 35.45] & 67.88 & 56.45 & 70.63 \\
MCU-CoT & 36.00 & [72.20, 73.45] & 51.79 & 65.37 \\
MCW-CoT & 42.77 & [81.87, 83.12] & 63.69 & 76.49 \\
MCU-CF & 41.92 & 68.86 & 66.47 & 75.40 \\
MCW-CF & 50.36 & 78.81 & 76.35 & 83.97 \\
\bottomrule\end{tabular}\end{adjustbox}\endgroup\end{center}

\textbf{Table 4b: Experiment 1 Auxiliaries --- Group C Endpoint Rates ($\ge 95\%$), Telic Interruption Sensitivity ($C-A$), and Group D \texttt{NLI-\allowbreak{}CoT} Counts.}

\begin{center}\begingroup\small\setlength{\tabcolsep}{4.5pt}\renewcommand{\arraystretch}{1.12}\begin{adjustbox}{max width=\textwidth}\begin{tabular}{lrrrr}\toprule
\textbf{Diagnostic Quantity} & \textbf{Direct} & \textbf{Direct-CoT} & \textbf{MCU} & \textbf{MCU-CoT} \\ \midrule
Group C Endpoint Rate (>=95\%, \%) & 88.75 & 26.25 & 1.25 & 0.00 \\
Telic Interruption (C-A, pp) / Group D NLI-CoT (T/F/U) & --- & 32.69 & 23.54 & 7/0/73 \\
\bottomrule\end{tabular}\end{adjustbox}\endgroup\end{center}

\textit{Note}: Continuous entries report percentages or worst-case Manski bounds \texttt{[lower, upper]}. Discrete \texttt{NLI-\allowbreak{}*} entries report \texttt{True/\allowbreak{}False/\allowbreak{}Unknown} counts (\texttt{E1} reports gold-label accuracy \texttt{\%}).

\begin{figure}[htbp]\centering
\includegraphics[width=0.98\textwidth]{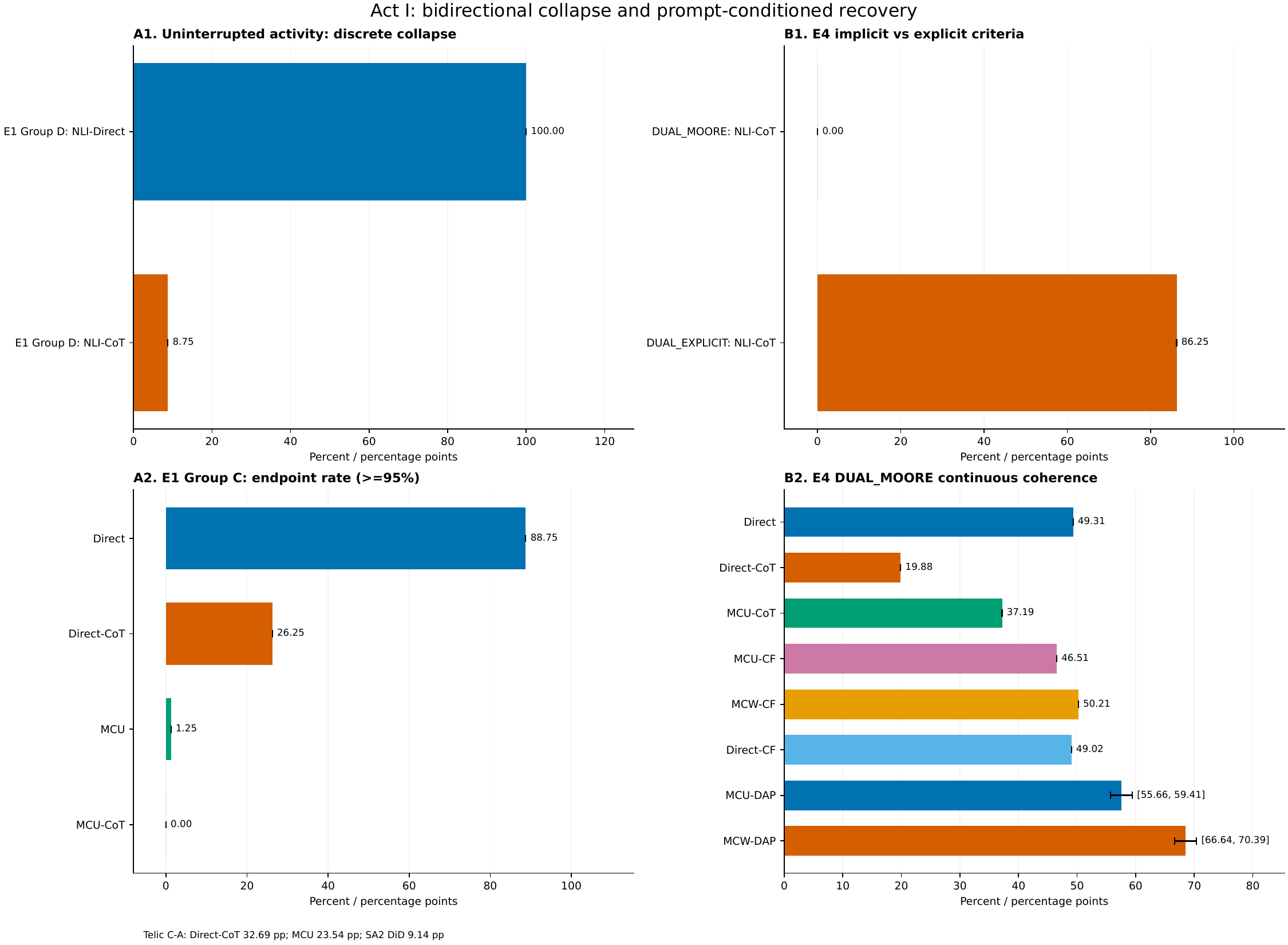}
\caption{Act I --- Single-chain lock-in and bidirectional collapse across English aspectual entailment (E1) and implicit Moorean pragmatics (E4).}
\end{figure}

\section{Act II: Cross-Linguistic Semantic Dissection and the Interruption-Anchored Realis Paradox}

Experiment 2 (\texttt{exp2-\allowbreak{}zh}, $80$ Mandarin four-slot clusters: $48$ telic, $32$ atelic; Table 5) crosses explicit interruption (\texttt{NI} vs. \texttt{INT}) with bare perfective \texttt{-\allowbreak{}le} (\texttt{NI\_\allowbreak{}BARE}, \texttt{INT\_\allowbreak{}BARE}, targeting event occurrence/initiation) versus explicit resultative completion compounds (\texttt{NI\_\allowbreak{}COMP}, \texttt{INT\_\allowbreak{}COMP}, targeting culmination). On uninterrupted telic predicates ($n=48$), \texttt{Direct} assigns \texttt{98.69\%} to \texttt{NI\_\allowbreak{}BARE} versus \texttt{77.13\%} to \texttt{NI\_\allowbreak{}COMP}, confirming \texttt{E2\_\allowbreak{}H1} (21.56 pp, Holm $p=$ 0.0008). Switching from \texttt{Direct} (\texttt{77.13\%}) to \texttt{MCU} (\texttt{[54.51\%, 56.60\%]}) on \texttt{NI\_\allowbreak{}COMP} reduces telic completion projection by \texttt{[}20.54\texttt{, }22.62\texttt{]} pp, confirming \texttt{E2\_\allowbreak{}H2} (Holm $p=$ 0.0008). On interrupted atelic activities ($n=32$), \texttt{Direct} sharply separates bare occurrence (\texttt{INT\_\allowbreak{}BARE} = \texttt{99.80\%}) from resultative completion (\texttt{INT\_\allowbreak{}COMP} = \texttt{11.41\%}), confirming \texttt{E2\_\allowbreak{}H3} (88.40 pp, Holm $p=$ 0.0008). Under \texttt{NLI-\allowbreak{}CoT}, uninterrupted atelic \texttt{NI\_\allowbreak{}BARE} exceeds \texttt{NI\_\allowbreak{}COMP} (\texttt{12.50\%} vs. \texttt{0.00\%} \texttt{True}), confirming \texttt{E2\_\allowbreak{}H4} (12.50 pp, one-sided Tango Holm $p=$ 0.0253, two-sided $98.75\%$ CI [-6.18, 34.72] pp).

Experiment 3 (\texttt{exp3-\allowbreak{}en}, $80$ English atelic four-slot clusters: $64$ \texttt{A\_\allowbreak{}pure}, $16$ \texttt{A\_\allowbreak{}plan}; Table 6a) holds progressive activity descriptions paired with \texttt{E1} while replacing minimal occurrence (\texttt{NI\_\allowbreak{}OCC}, \texttt{INT\_\allowbreak{}OCC}) with planned or durational completion (\texttt{NI\_\allowbreak{}PLAN}, \texttt{INT\_\allowbreak{}PLAN}). Under \texttt{Direct}, occurrence exceeds plan completion by 61.30 pp without interruption (\texttt{E3\_\allowbreak{}H1}: \texttt{96.99\%} vs. \texttt{35.69\%}, Holm $p=$ 0.0008) and by 81.86 pp under interruption (\texttt{E3\_\allowbreak{}H2}: \texttt{97.11\%} vs. \texttt{15.25\%}, Holm $p=$ 0.0008). Under \texttt{MCU-\allowbreak{}CoT}, \texttt{NI\_\allowbreak{}PLAN} (\texttt{21.44\%}) and \texttt{INT\_\allowbreak{}PLAN} (\texttt{20.38\%}) converge near a low floor (preregistered descriptive \texttt{E3\_\allowbreak{}H3} = 1.06 pp, primary descriptive $98.75\%$ CI [-0.31, 2.62] pp, companion descriptive $98.75\%$ CI [-0.28, 2.58] pp), while \texttt{E3\_\allowbreak{}H4} (\texttt{NLI-\allowbreak{}CoT} on \texttt{NI\_\allowbreak{}OCC} minus \texttt{NI\_\allowbreak{}PLAN} = 2.50 pp, Holm $p=$ 0.2152) does not reject $H_0$ because \texttt{NLI-\allowbreak{}CoT} collapses \texttt{NI\_\allowbreak{}OCC} itself to \texttt{97.50\%} \texttt{Unknown}.

Crucially, juxtaposing \texttt{E2} and \texttt{E3} reveals the cross-linguistic \textbf{Interruption-Anchored Realis Paradox} under \texttt{NLI-\allowbreak{}CoT} (Table 6b, Figure 2). When an ongoing atelic activity is presented without an interruption clause (\texttt{NI\_\allowbreak{}BARE} in Mandarin, \texttt{NI\_\allowbreak{}OCC} in English), \texttt{NLI-\allowbreak{}CoT} imports an unstated completion standard and defaults to \texttt{Unknown} (\texttt{22/\allowbreak{}32} \texttt{Unknown} and \texttt{6/\allowbreak{}32} \texttt{False} in Mandarin; \texttt{78/\allowbreak{}80} \texttt{Unknown} in English). Yet when the exact same activity is \textbf{explicitly interrupted} (\texttt{INT\_\allowbreak{}BARE} / \texttt{INT\_\allowbreak{}OCC}), the interruption clause pragmatically presupposes that the activity was underway prior to being stopped, causing \texttt{NLI-\allowbreak{}CoT} \texttt{True} judgments to surge from \texttt{12.50\%} (\texttt{4/\allowbreak{}32}) to \texttt{96.88\%} (\texttt{31/\allowbreak{}32}) in Mandarin and from \texttt{2.50\%} (\texttt{2/\allowbreak{}80}) to \texttt{38.75\%} (\texttt{31/\allowbreak{}80}) in English (\texttt{81.25\%} under \texttt{NLI-\allowbreak{}Direct}, \texttt{98.75\%} under \texttt{NLI-\allowbreak{}DAP}). By contrast, \texttt{MCU} (\texttt{78.15\%} on \texttt{NI\_\allowbreak{}OCC} vs. \texttt{78.04\%} on \texttt{INT\_\allowbreak{}OCC}) avoids this discrete \texttt{Unknown} collapse while maintaining a $>53\text{ pp}$ separation above planned completion (\texttt{24.19\%} on \texttt{NI\_\allowbreak{}PLAN}, \texttt{23.41\%} on \texttt{INT\_\allowbreak{}PLAN}).

\textbf{Table 5a: Experiment 2 (telic) --- Mandarin Telic Predicates (n = 48) (13 Prompt Arms / 17 Canonical Readout Rows).}

\begin{center}\begingroup\small\setlength{\tabcolsep}{4.5pt}\renewcommand{\arraystretch}{1.12}\begin{adjustbox}{max width=\textwidth}\begin{tabular}{lrrrr}\toprule
\textbf{Canonical Method / Readout} & \textbf{NI\_\allowbreak{}COMP} & \textbf{NI\_\allowbreak{}BARE} & \textbf{INT\_\allowbreak{}COMP} & \textbf{INT\_\allowbreak{}BARE} \\ \midrule
NLI-Direct & 0/0/48 & 45/0/3 & 2/16/30 & 41/1/6 \\
NLI-DAP & 0/0/48 & 3/0/45 & 1/8/39 & 16/0/32 \\
NLI-CoT & 0/0/48 & 4/4/40 & 1/39/8 & 24/8/16 \\
NLI-CF & 0/0/48 & 44/0/4 & 1/43/4 & 33/10/5 \\
Strict-Logician & [40.62, 40.62] & [78.12, 78.12] & [3.12, 3.12] & [85.42, 85.42] \\
Direct & [77.13, 77.13] & [98.69, 98.69] & [3.42, 3.42] & [84.76, 84.76] \\
Direct-DAP & [39.17, 39.17] & [53.54, 53.54] & [4.94, 4.94] & [18.12, 18.12] \\
Direct-CoT & [42.81, 42.81] & [47.60, 47.60] & [5.16, 5.16] & [56.14, 56.14] \\
Direct-CF & [67.92, 67.92] & [76.56, 76.56] & [4.69, 4.69] & [41.04, 41.04] \\
MCU & [54.51, 56.60] & [66.20, 66.20] & [19.40, 19.40] & [52.67, 56.84] \\
MCW & [74.15, 76.24] & [86.56, 86.56] & [13.97, 13.97] & [68.74, 72.91] \\
MCU-DAP & [37.73, 37.73] & [42.06, 42.06] & [20.86, 20.86] & [26.40, 30.56] \\
MCW-DAP & [49.64, 49.64] & [56.26, 56.26] & [14.29, 14.29] & [23.97, 28.14] \\
MCU-CoT & [38.00, 38.00] & [43.02, 43.02] & [17.31, 19.39] & [45.23, 47.31] \\
MCW-CoT & [50.66, 50.66] & [59.52, 59.52] & [12.90, 14.98] & [60.04, 62.13] \\
MCU-CF & [53.12, 53.12] & [58.52, 58.52] & [20.23, 20.23] & [39.58, 39.58] \\
MCW-CF & [69.19, 69.19] & [77.19, 77.19] & [14.54, 14.54] & [52.94, 52.94] \\
\bottomrule\end{tabular}\end{adjustbox}\endgroup\end{center}

\textit{Note}: Continuous entries report percentages or worst-case Manski bounds \texttt{[lower, upper]}. Discrete \texttt{NLI-\allowbreak{}*} entries report \texttt{True/\allowbreak{}False/\allowbreak{}Unknown} counts (\texttt{E1} reports gold-label accuracy \texttt{\%}).

\textbf{Table 5b: Experiment 2 (atelic) --- Mandarin Atelic Activities (n = 32) (13 Prompt Arms / 17 Canonical Readout Rows).}

\begin{center}\begingroup\small\setlength{\tabcolsep}{4.5pt}\renewcommand{\arraystretch}{1.12}\begin{adjustbox}{max width=\textwidth}\begin{tabular}{lrrrr}\toprule
\textbf{Canonical Method / Readout} & \textbf{NI\_\allowbreak{}COMP} & \textbf{NI\_\allowbreak{}BARE} & \textbf{INT\_\allowbreak{}COMP} & \textbf{INT\_\allowbreak{}BARE} \\ \midrule
NLI-Direct & 0/0/32 & 32/0/0 & 2/2/28 & 32/0/0 \\
NLI-DAP & 0/0/32 & 2/0/30 & 0/0/32 & 30/0/2 \\
NLI-CoT & 0/0/32 & 4/6/22 & 1/17/14 & 31/0/1 \\
NLI-CF & 0/0/32 & 31/0/1 & 0/16/16 & 32/0/0 \\
Strict-Logician & [46.88, 46.88] & [79.69, 79.69] & [28.12, 28.12] & [100.00, 100.00] \\
Direct & [79.69, 79.69] & [94.97, 94.97] & [11.41, 11.41] & [99.80, 99.80] \\
Direct-DAP & [48.59, 48.59] & [78.84, 78.84] & [14.53, 14.53] & [67.19, 67.19] \\
Direct-CoT & [43.91, 43.91] & [47.19, 47.19] & [24.06, 24.06] & [91.52, 91.52] \\
Direct-CF & [66.41, 66.41] & [79.84, 79.84] & [12.19, 12.19] & [72.02, 72.02] \\
MCU & [53.74, 53.74] & [58.99, 58.99] & [23.62, 29.87] & [77.90, 81.02] \\
MCW & [71.13, 71.13] & [80.01, 80.01] & [23.48, 29.73] & [88.51, 91.64] \\
MCU-DAP & [30.78, 30.78] & [46.58, 46.58] & [26.75, 26.75] & [43.97, 43.97] \\
MCW-DAP & [38.90, 38.90] & [62.02, 62.02] & [26.27, 26.27] & [54.05, 54.05] \\
MCU-CoT & [36.41, 36.41] & [45.66, 45.66] & [23.06, 29.31] & [73.12, 73.12] \\
MCW-CoT & [49.38, 49.38] & [59.97, 59.97] & [23.45, 29.70] & [88.04, 88.04] \\
MCU-CF & [50.25, 50.25] & [59.53, 59.53] & [27.22, 27.22] & [50.38, 50.38] \\
MCW-CF & [71.70, 71.70] & [84.65, 84.65] & [32.83, 32.83] & [72.62, 72.62] \\
\bottomrule\end{tabular}\end{adjustbox}\endgroup\end{center}

\textit{Note}: Continuous entries report percentages or worst-case Manski bounds \texttt{[lower, upper]}. Discrete \texttt{NLI-\allowbreak{}*} entries report \texttt{True/\allowbreak{}False/\allowbreak{}Unknown} counts (\texttt{E1} reports gold-label accuracy \texttt{\%}).

\textbf{Table 6a: Experiment 3 (all) --- English Atelic Occurrence vs. Planned Completion (n = 80) (13 Prompt Arms / 17 Canonical Readout Rows).}

\begin{center}\begingroup\small\setlength{\tabcolsep}{4.5pt}\renewcommand{\arraystretch}{1.12}\begin{adjustbox}{max width=\textwidth}\begin{tabular}{lrrrr}\toprule
\textbf{Canonical Method / Readout} & \textbf{NI\_\allowbreak{}OCC} & \textbf{NI\_\allowbreak{}PLAN} & \textbf{INT\_\allowbreak{}OCC} & \textbf{INT\_\allowbreak{}PLAN} \\ \midrule
NLI-Direct & 31/0/49 & 0/0/80 & 65/0/15 & 0/7/73 \\
NLI-DAP & 58/0/22 & 0/0/80 & 79/0/1 & 0/7/73 \\
NLI-CoT & 2/0/78 & 0/0/80 & 31/0/49 & 0/26/54 \\
NLI-CF & 0/0/80 & 0/0/80 & 2/0/78 & 0/28/52 \\
Strict-Logician & 60.62 & 14.79 & 76.88 & 7.50 \\
Direct & 96.99 & 35.69 & 97.11 & 15.25 \\
Direct-DAP & 90.55 & 53.31 & 94.19 & 29.88 \\
Direct-CoT & 94.05 & 42.81 & 96.04 & 15.62 \\
Direct-CF & 89.09 & 37.00 & 85.80 & 15.31 \\
MCU & 78.15 & 24.19 & 78.04 & 23.41 \\
MCW & 86.42 & 27.97 & 86.74 & 25.30 \\
MCU-DAP & 71.01 & 34.69 & 77.31 & [25.52, 29.28] \\
MCW-DAP & 78.81 & 39.42 & 84.89 & [26.83, 30.58] \\
MCU-CoT & [69.10, 70.35] & 21.44 & [79.62, 80.87] & 20.38 \\
MCW-CoT & [81.51, 82.76] & 24.26 & [87.99, 89.24] & 21.23 \\
MCU-CF & [69.69, 70.94] & 32.39 & 67.92 & 22.79 \\
MCW-CF & [79.38, 80.63] & 35.35 & 76.96 & 23.77 \\
\bottomrule\end{tabular}\end{adjustbox}\endgroup\end{center}

\textit{Note}: Continuous entries report percentages or worst-case Manski bounds \texttt{[lower, upper]}. Discrete \texttt{NLI-\allowbreak{}*} entries report \texttt{True/\allowbreak{}False/\allowbreak{}Unknown} counts (\texttt{E1} reports gold-label accuracy \texttt{\%}).

\textbf{Table 6b: Cross-Linguistic Interruption-Anchored Realis Paradox Under \texttt{NLI-\allowbreak{}CoT} (Exact \texttt{True} / \texttt{False} / \texttt{Unknown} Counts).}

\begin{center}\begingroup\small\setlength{\tabcolsep}{4.5pt}\renewcommand{\arraystretch}{1.12}\begin{adjustbox}{max width=\textwidth}\begin{tabular}{lrrrr}\toprule
\textbf{Experiment / Subgroup / Slot} & \textbf{Canonical Method} & \textbf{True} & \textbf{False} & \textbf{Unknown} \\ \midrule
E2 atelic NI\_\allowbreak{}BARE & NLI-CoT & 4 & 6 & 22 \\
E2 atelic INT\_\allowbreak{}BARE & NLI-CoT & 31 & 0 & 1 \\
E3 all NI\_\allowbreak{}OCC & NLI-CoT & 2 & 0 & 78 \\
E3 all INT\_\allowbreak{}OCC & NLI-CoT & 31 & 0 & 49 \\
\bottomrule\end{tabular}\end{adjustbox}\endgroup\end{center}

\begin{figure}[htbp]\centering
\includegraphics[width=0.98\textwidth]{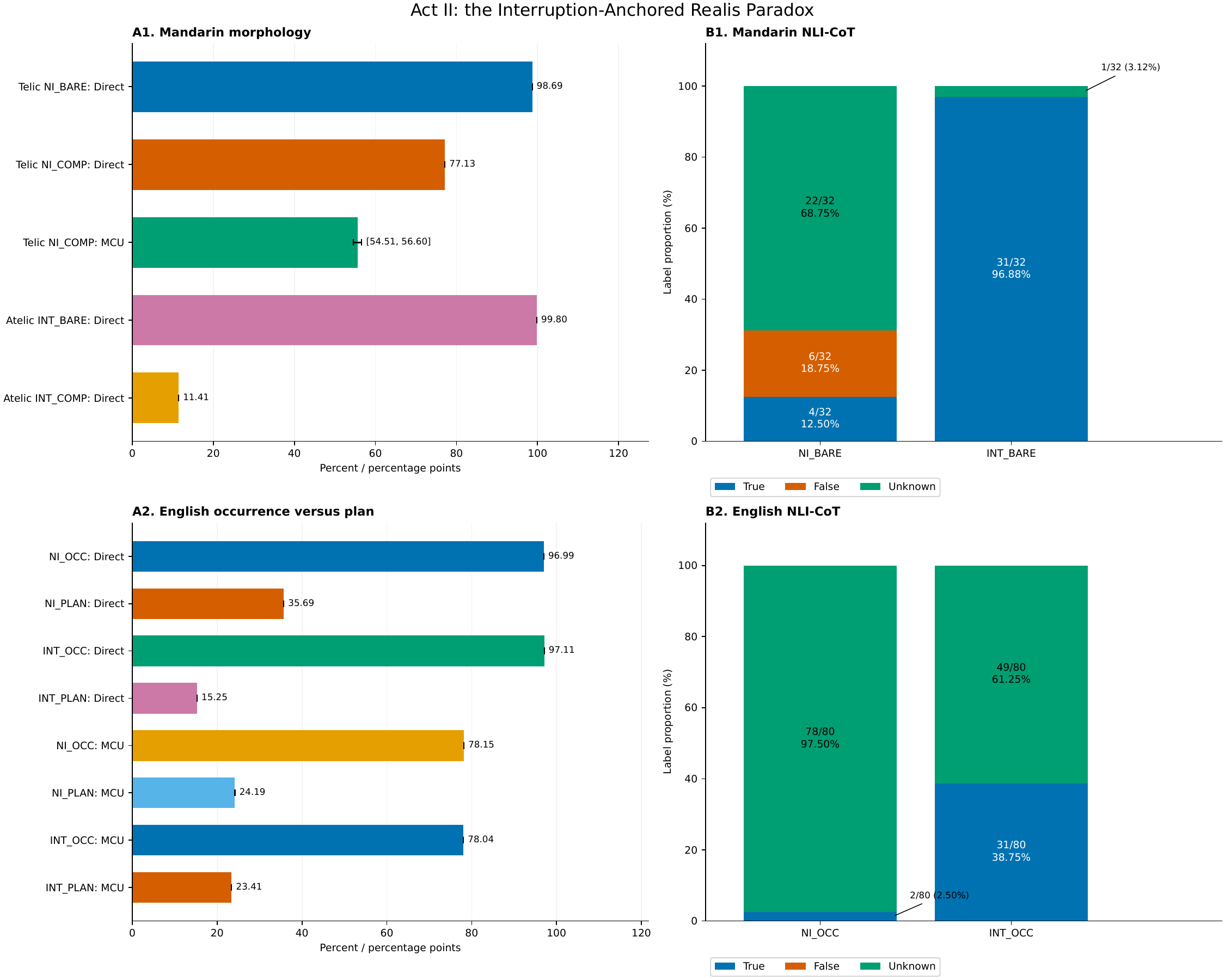}
\caption{Act II --- Cross-linguistic semantic dissection and the Interruption-Anchored Realis Paradox in Mandarin (E2) and English (E3).}
\end{figure}

\section{Act III: The Directional Reversal Test and Pragmatic Dual-Criteria Decomposition}

A natural skepticism regarding \texttt{MCV} in Acts I and II is that averaging five candidate estimates might merely pull extreme probabilities downward toward $50\%$. Experiment 4 (\texttt{exp4-\allowbreak{}moore}, Table 7a) directly tests this alternative via the \textbf{Directional Reversal Test} (\texttt{E4\_\allowbreak{}H1}). On implicit dual-criteria Moorean sentences (\texttt{DUAL\_\allowbreak{}MOORE}), where single-chain \texttt{Direct-\allowbreak{}CoT} drives perceived coherence down to \texttt{19.88\%}, \texttt{MCU-\allowbreak{}CF} shifts perceived coherence \textbf{upward by 26.64 pp} to \texttt{46.51\%} (primary $98.75\%$ CI [16.17, 36.43] pp, Holm $p=$ 0.0006; \texttt{MCW-\allowbreak{}CF} reaches \texttt{50.21\%}, a \texttt{+30.33 pp} upward shift). This positive reversal holds across all four pragmatic domains (\texttt{D1\_\allowbreak{}safety}: \texttt{28.95 pp}, \texttt{D2\_\allowbreak{}project}: \texttt{22.80 pp}, \texttt{D3\_\allowbreak{}medical}: \texttt{18.00 pp}, \texttt{D4\_\allowbreak{}institutional}: \texttt{36.80 pp}; Table 7b).

Because \texttt{E4\_\allowbreak{}H1} compares \texttt{MCU-\allowbreak{}CF} against \texttt{Direct-\allowbreak{}CoT}, it simultaneously alters both the prompt instructions (adding counterfactual dual-criteria guidance \texttt{CF}) and the output/aggregation protocol (\texttt{MCU} vs. single-point output). Therefore, \texttt{E4\_\allowbreak{}H1} alone neither rules out regression toward $50\%$ nor isolates multi-candidate aggregation as a standalone causal factor. To disentangle these components, Table 7d reports the preregistered matched-prompt companion contrasts:
1. \textbf{Counterfactual Prompt Effect at Fixed Single-Point Readout (\texttt{Direct-\allowbreak{}CF} minus \texttt{Direct-\allowbreak{}CoT})}: Holding the single-probability output bottleneck fixed, adding \texttt{CF} instructions raises \texttt{DUAL\_\allowbreak{}MOORE} coherence by 29.15 pp (\texttt{49.02\%} vs. \texttt{19.88\%}).
2. \textbf{Multi-Candidate Protocol Effect at Fixed CoT Instructions (\texttt{MCU-\allowbreak{}CoT} minus \texttt{Direct-\allowbreak{}CoT})}: Holding standard \texttt{CoT} instructions fixed without \texttt{CF} hints, switching from single-point \texttt{Direct-\allowbreak{}CoT} to five-candidate \texttt{MCU-\allowbreak{}CoT} raises \texttt{DUAL\_\allowbreak{}MOORE} coherence by 17.31 pp (\texttt{37.19\%} vs. \texttt{19.88\%}; and \texttt{MCW-\allowbreak{}CoT} reaches \texttt{42.22\%}).
3. \textbf{Above-Midpoint Separation Under Discourse-Anchor Paraphrase (\texttt{MCU-\allowbreak{}DAP} and \texttt{MCW-\allowbreak{}DAP})}: On the \texttt{DAP} arm, \texttt{MCU-\allowbreak{}DAP} (\texttt{[55.66\%, 59.41\%]}) and \texttt{MCW-\allowbreak{}DAP} (\texttt{[66.64\%, 70.39\%]}) place \texttt{DUAL\_\allowbreak{}MOORE} strictly above $50\%$ while keeping true single-standard contradictions (\texttt{PURE\_\allowbreak{}MOORE}) low (\texttt{[21.75\%, 23.00\%]} on \texttt{MCU-\allowbreak{}DAP}), refuting a universal midpoint-compression account.

Reasoning-trace markers on \texttt{DUAL\_\allowbreak{}MOORE} (Table 7c, Figure 3B) align with this decomposition: textbook single-standard contradiction reflexes drop from \texttt{68.75\%} (\texttt{55/\allowbreak{}80}) under \texttt{NLI-\allowbreak{}CoT} and \texttt{53.75\%} (\texttt{43/\allowbreak{}80}) under \texttt{Direct-\allowbreak{}CoT} to \texttt{41.25\%} (\texttt{33/\allowbreak{}80}) under \texttt{MCV-\allowbreak{}CoT}, \texttt{47.50\%} (\texttt{38/\allowbreak{}80}) under \texttt{MCV-\allowbreak{}CF}, and \texttt{22.50\%} (\texttt{18/\allowbreak{}80}) under \texttt{NLI-\allowbreak{}CF}, while explicit detection of dual criteria or polysemy rises from \texttt{0.00\%} (\texttt{0/\allowbreak{}80}) under \texttt{NLI-\allowbreak{}CoT} and \texttt{3.75\%} (\texttt{3/\allowbreak{}80}) under \texttt{Direct-\allowbreak{}CoT} to \texttt{47.50\%} (\texttt{38/\allowbreak{}80}) under \texttt{Direct-\allowbreak{}CF}, \texttt{52.50\%} (\texttt{42/\allowbreak{}80}) under \texttt{MCV-\allowbreak{}CF}, and \texttt{92.50\%} (\texttt{74/\allowbreak{}80}) under \texttt{NLI-\allowbreak{}CF}.

At the same time, slot \texttt{E4\_\allowbreak{}H2} (\texttt{DUAL\_\allowbreak{}MOORE} minus \texttt{PURE\_\allowbreak{}MOORE} under unguided zero-shot \texttt{MCU}) highlights an important boundary condition (Table 7d): without reasoning or structural guidance, \texttt{MCV-\allowbreak{}Direct} incurs 12 missing format units on \texttt{DUAL\_\allowbreak{}MOORE} (\texttt{68/\allowbreak{}80} valid) and 3 on \texttt{PURE\_\allowbreak{}MOORE} (\texttt{77/\allowbreak{}80} valid), totaling 15 missing units across 14 clusters. Under worst-case Manski bounds, \texttt{E4\_\allowbreak{}H2} spans \texttt{[}-2.05\texttt{, }16.70\texttt{]} pp (Holm $p=$ 1.0000, not rejected), even though the 66 complete pairs yield \texttt{+}5.02\texttt{ pp} and the available marginal difference is \texttt{+}6.01\texttt{ pp} (\texttt{36.32\%} vs. \texttt{30.31\%}).

\textbf{Table 7a: Experiment 4 (all) --- English Moorean Pragmatics \& Implicit Dual Criteria (n = 80) (13 Prompt Arms / 17 Canonical Readout Rows).}

\begin{center}\begingroup\small\setlength{\tabcolsep}{4.5pt}\renewcommand{\arraystretch}{1.12}\begin{adjustbox}{max width=\textwidth}\begin{tabular}{lrrrr}\toprule
\textbf{Canonical Method / Readout} & \textbf{PURE\_\allowbreak{}MOORE} & \textbf{DUAL\_\allowbreak{}MOORE} & \textbf{DUAL\_\allowbreak{}EXPLICIT} & \textbf{PURE\_\allowbreak{}READ} \\ \midrule
NLI-Direct & 2/2/76 & 9/4/67 & 68/0/12 & 25/0/55 \\
NLI-DAP & 80/0/0 & 80/0/0 & 80/0/0 & 80/0/0 \\
NLI-CoT & 0/79/1 & 0/80/0 & 69/9/2 & 48/18/14 \\
NLI-CF & 0/5/75 & 2/2/76 & 64/0/16 & 20/0/60 \\
Strict-Logician & 0.00 & 0.00 & 84.06 & 28.12 \\
Direct & 15.31 & 49.31 & 97.69 & 80.12 \\
Direct-DAP & 1.88 & 69.62 & 95.69 & 95.94 \\
Direct-CoT & 10.62 & 19.88 & 97.50 & 52.31 \\
Direct-CF & 37.92 & 49.02 & 75.36 & 53.55 \\
MCU & [29.18, 32.92] & [30.88, 45.88] & [80.96, 82.21] & [61.90, 64.40] \\
MCW & [33.46, 37.21] & [37.46, 52.46] & [87.92, 89.17] & [69.77, 72.27] \\
MCU-DAP & [21.75, 23.00] & [55.66, 59.41] & 74.18 & 71.40 \\
MCW-DAP & [20.53, 21.78] & [66.64, 70.39] & 87.05 & 85.91 \\
MCU-CoT & 25.84 & 37.19 & 72.75 & 37.80 \\
MCW-CoT & 27.65 & 42.23 & 80.24 & 41.75 \\
MCU-CF & 35.18 & 46.51 & 63.19 & 49.51 \\
MCW-CF & 35.73 & 50.21 & 69.62 & 54.12 \\
\bottomrule\end{tabular}\end{adjustbox}\endgroup\end{center}

\textit{Note}: Continuous entries report percentages or worst-case Manski bounds \texttt{[lower, upper]}. Discrete \texttt{NLI-\allowbreak{}*} entries report \texttt{True/\allowbreak{}False/\allowbreak{}Unknown} counts (\texttt{E1} reports gold-label accuracy \texttt{\%}).

\textbf{Table 7b: Experiment 4 Domain Breakdown for \texttt{E4\_\allowbreak{}H1} (\texttt{DUAL\_\allowbreak{}MOORE}: \texttt{MCU-\allowbreak{}CF} minus \texttt{Direct-\allowbreak{}CoT}, $n=20$ per domain).}

\begin{center}\begingroup\small\setlength{\tabcolsep}{4.5pt}\renewcommand{\arraystretch}{1.12}\begin{adjustbox}{max width=\textwidth}\begin{tabular}{llll}\toprule
\textbf{Pragmatic Domain Slice} & \textbf{Clusters (n)} & \textbf{Canonical Contrast} & \textbf{Directional Reversal Shift (pp)} \\ \midrule
D1\_\allowbreak{}safety & 20 & DUAL\_\allowbreak{}MOORE: MCU-CF - Direct-CoT & 28.95 \\
D2\_\allowbreak{}project & 20 & DUAL\_\allowbreak{}MOORE: MCU-CF - Direct-CoT & 22.80 \\
D3\_\allowbreak{}medical & 20 & DUAL\_\allowbreak{}MOORE: MCU-CF - Direct-CoT & 18.00 \\
D4\_\allowbreak{}institutional & 20 & DUAL\_\allowbreak{}MOORE: MCU-CF - Direct-CoT & 36.80 \\
\bottomrule\end{tabular}\end{adjustbox}\endgroup\end{center}

\textbf{Table 7c: Experiment 4 Reasoning-Trace Mechanism Audit on \texttt{DUAL\_\allowbreak{}MOORE} ($n=80$).}

\begin{center}\begingroup\small\setlength{\tabcolsep}{4.5pt}\renewcommand{\arraystretch}{1.12}\begin{adjustbox}{max width=\textwidth}\begin{tabular}{lllll}\toprule
\textbf{Canonical Reasoning Method} & \textbf{Textbook Single-Standard Reflex Count} & \textbf{Textbook Reflex Rate (\%)} & \textbf{Dual-Criteria / Polysemy Count} & \textbf{Dual-Criteria Detection Rate (\%)} \\ \midrule
NLI-CoT & 55 & 68.75 & 0 & 0.00 \\
Direct-CoT & 43 & 53.75 & 3 & 3.75 \\
MCV-CoT & 33 & 41.25 & 3 & 3.75 \\
Direct-CF & 47 & 58.75 & 38 & 47.50 \\
MCV-CF & 38 & 47.50 & 42 & 52.50 \\
NLI-CF & 18 & 22.50 & 74 & 92.50 \\
\bottomrule\end{tabular}\end{adjustbox}\endgroup\end{center}

\textbf{Table 7d: Experiment 4 Matched-Prompt Companion Decompositions and \texttt{E4\_\allowbreak{}H2} Missingness Diagnostics.}

\begin{center}\begingroup\small\setlength{\tabcolsep}{4.5pt}\renewcommand{\arraystretch}{1.12}\begin{adjustbox}{max width=\textwidth}\begin{tabular}{llll}\toprule
\textbf{Decomposition Family / Diagnostic} & \textbf{Canonical Contrast / Quantity} & \textbf{Lower Bound / Value} & \textbf{Upper Bound / Companion} \\ \midrule
E4\_\allowbreak{}H1\_\allowbreak{}decompositions\_\allowbreak{}on\_\allowbreak{}DUAL\_\allowbreak{}MOORE & DUAL\_\allowbreak{}MOORE: MCU - Direct-CoT & 11.00 & 26.00 \\
E4\_\allowbreak{}H1\_\allowbreak{}decompositions\_\allowbreak{}on\_\allowbreak{}DUAL\_\allowbreak{}MOORE & DUAL\_\allowbreak{}MOORE: MCU-CoT - Direct-CoT & 17.31 & 17.31 \\
E4\_\allowbreak{}H1\_\allowbreak{}decompositions\_\allowbreak{}on\_\allowbreak{}DUAL\_\allowbreak{}MOORE & DUAL\_\allowbreak{}MOORE: Direct-CF - Direct-CoT & 29.15 & 29.15 \\
E4\_\allowbreak{}H1\_\allowbreak{}decompositions\_\allowbreak{}on\_\allowbreak{}DUAL\_\allowbreak{}MOORE & DUAL\_\allowbreak{}MOORE: MCW-CF - Direct-CoT & 30.33 & 30.33 \\
E4\_\allowbreak{}H2\_\allowbreak{}decompositions\_\allowbreak{}DUAL\_\allowbreak{}MOORE\_\allowbreak{}minus\_\allowbreak{}PURE\_\allowbreak{}MOORE & DUAL\_\allowbreak{}MOORE - PURE\_\allowbreak{}MOORE (MCU-CF) & 11.34 & 11.34 \\
E4\_\allowbreak{}H2\_\allowbreak{}decompositions\_\allowbreak{}DUAL\_\allowbreak{}MOORE\_\allowbreak{}minus\_\allowbreak{}PURE\_\allowbreak{}MOORE & DUAL\_\allowbreak{}MOORE - PURE\_\allowbreak{}MOORE (Direct) & 34.00 & 34.00 \\
E4\_\allowbreak{}H2\_\allowbreak{}decompositions\_\allowbreak{}DUAL\_\allowbreak{}MOORE\_\allowbreak{}minus\_\allowbreak{}PURE\_\allowbreak{}MOORE & DUAL\_\allowbreak{}MOORE - PURE\_\allowbreak{}MOORE (Direct-CoT) & 9.25 & 9.25 \\
E4\_\allowbreak{}H2\_\allowbreak{}decompositions\_\allowbreak{}DUAL\_\allowbreak{}MOORE\_\allowbreak{}minus\_\allowbreak{}PURE\_\allowbreak{}MOORE & DUAL\_\allowbreak{}MOORE - PURE\_\allowbreak{}MOORE (MCW) & 0.25 & 19.00 \\
E4\_\allowbreak{}H3\_\allowbreak{}decompositions\_\allowbreak{}DUAL\_\allowbreak{}EXPLICIT\_\allowbreak{}minus\_\allowbreak{}DUAL\_\allowbreak{}MOORE & DUAL\_\allowbreak{}EXPLICIT - DUAL\_\allowbreak{}MOORE (Direct-CoT) & 77.62 & 77.62 \\
E4\_\allowbreak{}H3\_\allowbreak{}decompositions\_\allowbreak{}DUAL\_\allowbreak{}EXPLICIT\_\allowbreak{}minus\_\allowbreak{}DUAL\_\allowbreak{}MOORE & DUAL\_\allowbreak{}EXPLICIT - DUAL\_\allowbreak{}MOORE (MCU) & 35.09 & 51.34 \\
E4\_\allowbreak{}H3\_\allowbreak{}decompositions\_\allowbreak{}DUAL\_\allowbreak{}EXPLICIT\_\allowbreak{}minus\_\allowbreak{}DUAL\_\allowbreak{}MOORE & DUAL\_\allowbreak{}EXPLICIT - DUAL\_\allowbreak{}MOORE (MCU-CF) & 16.68 & 16.68 \\
E4\_\allowbreak{}H4\_\allowbreak{}decompositions\_\allowbreak{}on\_\allowbreak{}L\_\allowbreak{}true\_\allowbreak{}pct & PURE\_\allowbreak{}READ - DUAL\_\allowbreak{}MOORE (NLI-CoT True \%) & 60.00 & 60.00 \\
E4\_\allowbreak{}H4\_\allowbreak{}decompositions\_\allowbreak{}on\_\allowbreak{}L\_\allowbreak{}true\_\allowbreak{}pct & DUAL\_\allowbreak{}MOORE: NLI-CF - NLI-CoT (True \%) & 2.50 & 2.50 \\
E4\_\allowbreak{}H4\_\allowbreak{}decompositions\_\allowbreak{}on\_\allowbreak{}L\_\allowbreak{}true\_\allowbreak{}pct & DUAL\_\allowbreak{}EXPLICIT - DUAL\_\allowbreak{}MOORE (NLI-Direct True \%) & 73.75 & 73.75 \\
E4\_\allowbreak{}H4\_\allowbreak{}decompositions\_\allowbreak{}on\_\allowbreak{}L\_\allowbreak{}true\_\allowbreak{}pct & DUAL\_\allowbreak{}EXPLICIT - DUAL\_\allowbreak{}MOORE (NLI-CF True \%) & 77.50 & 77.50 \\
E4\_\allowbreak{}H2 Missingness (MCV-Direct) & DUAL\_\allowbreak{}MOORE valid / missing; PURE\_\allowbreak{}MOORE valid / missing & 68 / 12 & 77 / 3 \\
E4\_\allowbreak{}H2 Complete-Pair \& Marginal Summary & Missing units / clusters; Complete pairs (n) / mean diff / marginal diff & 15 / 14 & 66 pairs: 5.02 pp (marginal 6.01 pp) \\
\bottomrule\end{tabular}\end{adjustbox}\endgroup\end{center}

\begin{figure}[htbp]\centering
\includegraphics[width=0.98\textwidth]{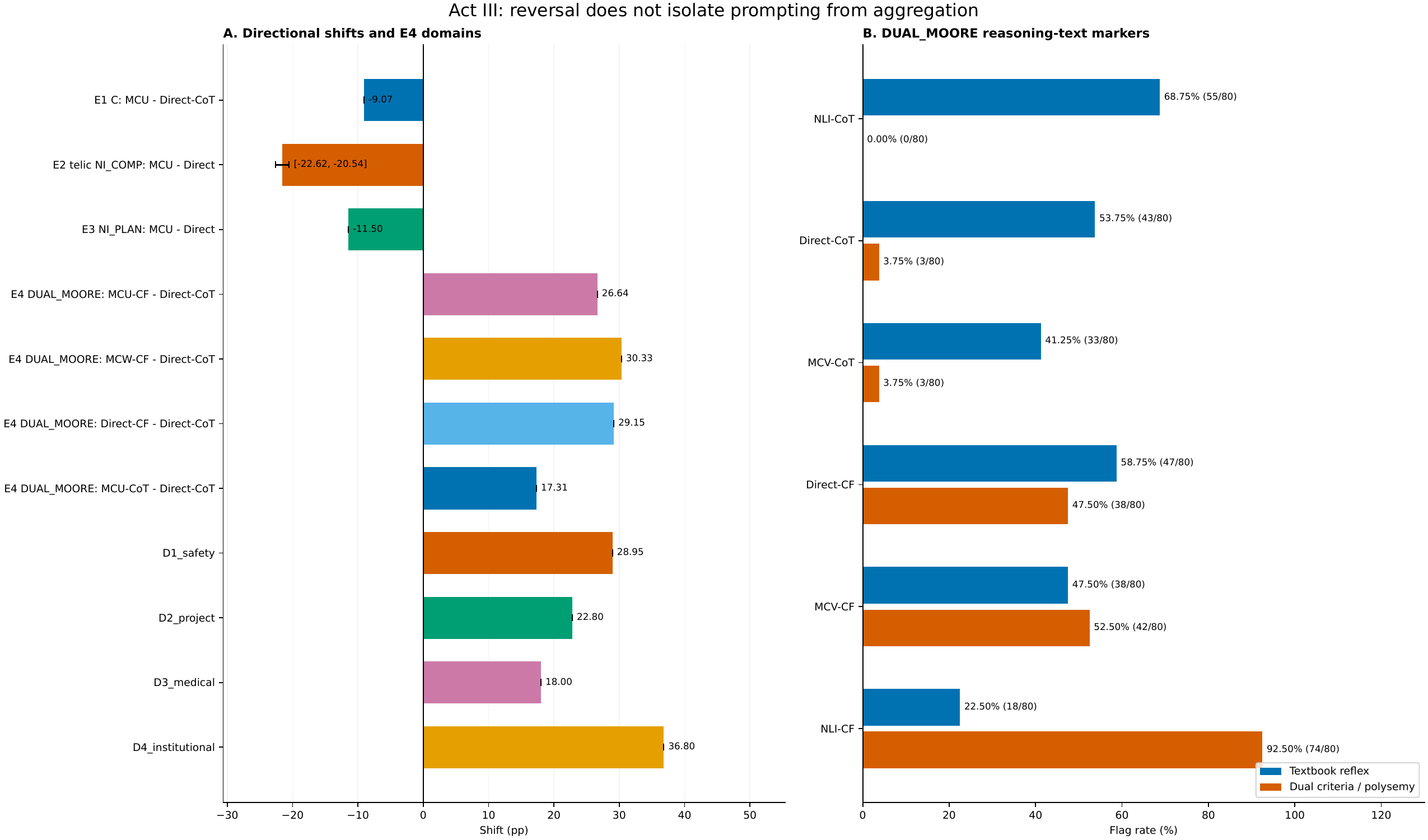}
\caption{Act III --- The Directional Reversal Test on implicit Moorean assertions (E4), domain breakdown, and reasoning-trace mechanism shift.}
\end{figure}

\section{Act IV: Self-Weighting Shift (\texttt{MCW} minus \texttt{MCU}), Format Ablation, and Extension Design}

Within every \texttt{MCV} call, comparing \texttt{MCW} against \texttt{MCU} isolates the effect of the model's self-reported candidate weights $\tilde{w}_k$ while holding the five generated probability estimates $p_1,\dots,p_5$ identical. Table 8 and Figure 4A show that \texttt{MCW} minus \texttt{MCU} is consistently positive across zero-missing benchmark slices---for example, \texttt{+7.35 pp} under \texttt{MCV-\allowbreak{}Direct} and \texttt{+11.90 pp} under \texttt{MCV-\allowbreak{}CoT} in \texttt{E1} Group C, \texttt{+}16.07\texttt{ pp} on telic \texttt{NI\_\allowbreak{}COMP} under \texttt{MCV-\allowbreak{}CF} in \texttt{E2}, \texttt{+8.27 pp} on \texttt{NI\_\allowbreak{}OCC} under \texttt{MCV-\allowbreak{}Direct} in \texttt{E3}, and \texttt{+3.70 pp} on \texttt{DUAL\_\allowbreak{}MOORE} under \texttt{MCV-\allowbreak{}CF} in \texttt{E4}. Because self-reported weights assign higher mass to dominant, high-confidence candidates, \texttt{MCW} shifts readouts back toward the model's primary mode, whereas equal-weight \texttt{MCU} preserves minority-candidate counterweight when guarding against single-standard over-projection.

To test how \texttt{MCV} output formatting and weight elicitation interact with token-level uncertainty, we conducted a $1{,}800$-call OpenAI ablation on \texttt{gpt-\allowbreak{}4o-\allowbreak{}mini} (\texttt{900} calls) and \texttt{gpt-\allowbreak{}4.1-\allowbreak{}mini} (\texttt{900} calls) across $150$ development contexts per variant at \texttt{temperature=\allowbreak{}0.6}, \texttt{top\_\allowbreak{}p=\allowbreak{}1.0}, and \texttt{top\_\allowbreak{}logprobs=\allowbreak{}20} (Table 9, Figure 4B). We compare six variants: \texttt{Direct (D0)}, \texttt{Direct-\allowbreak{}CoT (D\_\allowbreak{}CoT)}, standard sum-to-one weighted \texttt{MCV-\allowbreak{}Weighted (V0)}, free-weight \texttt{MCV-\allowbreak{}FreeWeight (V1)}, prompt-level unweighted \texttt{MCV-\allowbreak{}UnweightedPrompt (V2)} (which omits weights from the prompt), and strict \texttt{MCV-\allowbreak{}JSON (V3)}. All $1{,}800$ calls parsed at \texttt{100\%}. Removing weights from the prompt in \texttt{MCV-\allowbreak{}UnweightedPrompt (V2)} shrinks within-call candidate dispersion (\texttt{mean\_\allowbreak{}within\_\allowbreak{}call\_\allowbreak{}sd} drops from \texttt{15.23 pp} on \texttt{V0} to \texttt{6.59 pp} on \texttt{V2} in \texttt{gpt-\allowbreak{}4o-\allowbreak{}mini}, and from \texttt{22.26 pp} to \texttt{7.85 pp} in \texttt{gpt-\allowbreak{}4.1-\allowbreak{}mini}) and halves binary entropy gain $\Delta H_2$, confirming that eliciting candidate weights in the prompt scaffolds broader candidate separation even when downstream aggregation uses equal-weight \texttt{MCU}. Conversely, forcing strict JSON in \texttt{MCV-\allowbreak{}JSON (V3)} severely degrades semantic discrimination despite high token entropy.

\textbf{Table 8: Within-Call Self-Weighting Shift (\texttt{MCW} minus \texttt{MCU}) Across E1--E4 Benchmark Slices.}

\begin{center}\begingroup\small\setlength{\tabcolsep}{4.5pt}\renewcommand{\arraystretch}{1.12}\begin{adjustbox}{max width=\textwidth}\begin{tabular}{llll}\toprule
\textbf{Experiment \& Subgroup} & \textbf{Construction Slot} & \textbf{Canonical MCV Prompt} & \textbf{MCW - MCU Shift / Bounds (pp)} \\ \midrule
E1 Group C (Telic Uninterrupted) & Group C & MCV-Direct & 7.35 \\
E1 Group C (Telic Uninterrupted) & Group C & MCV-DAP & 5.80 \\
E1 Group C (Telic Uninterrupted) & Group C & MCV-CoT & 11.90 \\
E1 Group C (Telic Uninterrupted) & Group C & MCV-CF & 9.88 \\
E2 Mandarin (telic) & NI\_\allowbreak{}COMP & MCV-CF & [16.07, 16.07] \\
E2 Mandarin (telic) & NI\_\allowbreak{}BARE & MCV-CF & [18.67, 18.67] \\
E2 Mandarin (atelic) & NI\_\allowbreak{}COMP & MCV-CF & [21.45, 21.45] \\
E2 Mandarin (atelic) & NI\_\allowbreak{}BARE & MCV-CF & [25.11, 25.11] \\
E2 Mandarin (A\_\allowbreak{}pure) & NI\_\allowbreak{}COMP & MCV-CF & [18.91, 18.91] \\
E2 Mandarin (A\_\allowbreak{}pure) & NI\_\allowbreak{}BARE & MCV-CF & [24.56, 24.56] \\
E2 Mandarin (A\_\allowbreak{}plan) & NI\_\allowbreak{}COMP & MCV-CF & [23.98, 23.98] \\
E2 Mandarin (A\_\allowbreak{}plan) & NI\_\allowbreak{}BARE & MCV-CF & [25.67, 25.67] \\
E3 all & NI\_\allowbreak{}OCC & MCV-Direct & 8.27 \\
E3 all & NI\_\allowbreak{}PLAN & MCV-Direct & 3.78 \\
E3 all & INT\_\allowbreak{}OCC & MCV-Direct & 8.70 \\
E3 all & INT\_\allowbreak{}PLAN & MCV-Direct & 1.89 \\
E4 all & PURE\_\allowbreak{}MOORE & MCV-CF & 0.56 \\
E4 all & DUAL\_\allowbreak{}MOORE & MCV-CF & 3.70 \\
E4 all & DUAL\_\allowbreak{}EXPLICIT & MCV-CF & 6.43 \\
E4 all & PURE\_\allowbreak{}READ & MCV-CF & 4.61 \\
E4 all & PURE\_\allowbreak{}MOORE & MCV-CoT & 1.81 \\
E4 all & DUAL\_\allowbreak{}MOORE & MCV-CoT & 5.04 \\
E4 all & DUAL\_\allowbreak{}EXPLICIT & MCV-CoT & 7.49 \\
E4 all & PURE\_\allowbreak{}READ & MCV-CoT & 3.95 \\
\bottomrule\end{tabular}\end{adjustbox}\endgroup\end{center}

\textbf{Table 9: OpenAI 1,800-Call Format and Token-Entropy Ablation Summary (\texttt{temperature=\allowbreak{}0.6}, \texttt{top\_\allowbreak{}logprobs=\allowbreak{}20}).}

\begin{center}\begingroup\small\setlength{\tabcolsep}{4.5pt}\renewcommand{\arraystretch}{1.12}\begin{adjustbox}{max width=\textwidth}\begin{tabular}{lllllll}\toprule
\textbf{Model (Total Calls)} & \textbf{Canonical Variant} & \textbf{Parsed / Non-Copy Units} & \textbf{Semantic Strength H1 (pp)} & \textbf{First Numeric-Token Entropy (Non-Copy, bits)} & \textbf{Within-Call Candidate SD (pp)} & \textbf{Binary Entropy Gain vs. Direct (bits)} \\ \midrule
gpt-4o-mini (900) & Direct (D0) & 150 / 133 & 43.17 & 0.765 & 0.00 & 0.000 \\
gpt-4o-mini (900) & Direct-CoT (D\_\allowbreak{}CoT) & 150 / 146 & 39.33 & 0.903 & 0.00 & 0.064 \\
gpt-4o-mini (900) & MCV-Weighted (V0) & 150 / 137 & 26.49 & 1.260 & 15.23 & 0.211 \\
gpt-4o-mini (900) & MCV-FreeWeight (V1) & 150 / 150 & 24.48 & 1.454 & 15.37 & 0.224 \\
gpt-4o-mini (900) & MCV-UnweightedPrompt (V2) & 150 / 150 & 37.49 & 1.219 & 6.59 & 0.097 \\
gpt-4o-mini (900) & MCV-JSON (V3) & 150 / 150 & 13.40 & 1.400 & 19.70 & 0.303 \\
gpt-4.1-mini (900) & Direct (D0) & 150 / 136 & 40.95 & 0.648 & 0.00 & 0.000 \\
gpt-4.1-mini (900) & Direct-CoT (D\_\allowbreak{}CoT) & 150 / 149 & 42.33 & 0.683 & 0.00 & 0.060 \\
gpt-4.1-mini (900) & MCV-Weighted (V0) & 150 / 25 & 23.91 & 0.720 & 22.26 & 0.239 \\
gpt-4.1-mini (900) & MCV-FreeWeight (V1) & 150 / 27 & 26.45 & 0.955 & 19.13 & 0.215 \\
gpt-4.1-mini (900) & MCV-UnweightedPrompt (V2) & 150 / 110 & 36.52 & 1.079 & 7.85 & 0.101 \\
gpt-4.1-mini (900) & MCV-JSON (V3) & 150 / 150 & 23.71 & 1.325 & 20.46 & 0.263 \\
\bottomrule\end{tabular}\end{adjustbox}\endgroup\end{center}

The P16.7 extension comprises 4240 planned/collected units (4222 valid, 18 format-missing, 0 transport-missing) across five external models at 20 clusters per experiment, stage="dev", with less than 2\% missing in each completed slice; OpenAI gpt-4o-mini at top\_\allowbreak{}logprobs=20 has 1,000/1,000 valid units and zero missing. These dev slices use primary hybrid\_\allowbreak{}pct\_\allowbreak{}tango families, worst-case Manski bounds retaining missing units, simultaneous 98.75\% CIs and within-experiment Holm adjustment (B=5,000; locked seeds E2=2026110302, E3=2026110402, E4=2026110602), distinct from the Qwen3-8B formal S7 primary families. Table 10 reports model-specific bounds, Holm p-values, rejection counts and diagnostics.

\textbf{Table 10: P16.7 Cross-Model Confirmatory Dev Slices (\texttt{stage =\allowbreak{} dev}, 20 clusters/experiment).}

\begin{center}\begingroup\small\setlength{\tabcolsep}{4.5pt}\renewcommand{\arraystretch}{1.12}\begin{adjustbox}{max width=\textwidth}\begin{tabular}{lrr}\toprule
\textbf{Model} & \textbf{ok / planned} & \textbf{Missing \% (format / transport)} \\ \midrule
OpenAI 4o-mini & 1000/1000 & 0.00 (0/0) \\
Gemma 31B & 1000/1000 & 0.00 (0/0) \\
Nemotron 4B & 996/1000 & 0.40 (4/0) \\
Nemotron 30B & 986/1000 & 1.40 (14/0) \\
Groq OSS 120B & 240/240 & 0.00 (0/0) \\
\bottomrule\end{tabular}\end{adjustbox}\endgroup\end{center}

\begin{center}\begingroup\small\setlength{\tabcolsep}{4.5pt}\renewcommand{\arraystretch}{1.12}\begin{adjustbox}{max width=\textwidth}\begin{tabular}{lrrrrr}\toprule
\textbf{Slot: bounds pp; Holm p} & \textbf{OpenAI 4o-mini} & \textbf{Gemma 31B} & \textbf{Nemotron 4B} & \textbf{Nemotron 30B} & \textbf{Groq OSS 120B} \\ \midrule
E2\_\allowbreak{}H1 & 40.42; 0.0008 & 24.50; 0.0008 & 32.75; 0.0008 & 27.52; 0.0008 & 45.32; 0.0008 \\
E2\_\allowbreak{}H2 & 15.00; 0.0008 & 5.27; 0.2827 & -1.08; 1.0000 & [6.48, 14.81]; 0.2374 & -1.50; 1.0000 \\
E2\_\allowbreak{}H3 & 71.88; 0.0008 & 98.00; 0.0008 & 86.88; 0.0008 & 59.31; 0.0012 & 94.75; 0.0008 \\
E2\_\allowbreak{}H4 & 0.00; 1.0000 & 25.00; 0.2827 & 12.50; 0.5516 & 37.50; 0.1122 & 100.00; 0.0078 \\
E3\_\allowbreak{}H1 & 56.45; 0.0008 & 38.70; 0.0006 & 44.95; 0.0008 & 55.45; 0.0008 & uncollected \\
E3\_\allowbreak{}H2 & 78.50; 0.0008 & 85.30; 0.0006 & 75.40; 0.0008 & 80.25; 0.0008 & uncollected \\
E3\_\allowbreak{}H4 & 5.00; 0.5385 & 100.00; 3.81470e-06 & 5.00; 0.5385 & 50.00; 0.0009 & uncollected \\
E4\_\allowbreak{}H1 & 46.55; 0.0008 & 10.70; 0.1120 & -19.61; 1.0000 & -20.65; 1.0000 & uncollected \\
E4\_\allowbreak{}H2 & 29.20; 0.0008 & 18.38; 0.0612 & -8.47; 1.0000 & [-0.80, 14.20]; 1.0000 & uncollected \\
E4\_\allowbreak{}H3 & 72.25; 0.0008 & 37.75; 0.0006 & 19.00; 0.0128 & [0.00, 5.00]; 1.0000 & uncollected \\
E4\_\allowbreak{}H4 & 55.00; 0.0008 & 90.00; 7.57755e-06 & [10.00, 15.00]; 0.3307 & 75.00; 9.90973e-05 & uncollected \\
\bottomrule\end{tabular}\end{adjustbox}\endgroup\end{center}

\begin{center}\begingroup\small\setlength{\tabcolsep}{4.5pt}\renewcommand{\arraystretch}{1.12}\begin{adjustbox}{max width=\textwidth}\begin{tabular}{lrrrrr}\toprule
\textbf{Diagnostic: point bounds pp} & \textbf{OpenAI 4o-mini} & \textbf{Gemma 31B} & \textbf{Nemotron 4B} & \textbf{Nemotron 30B} & \textbf{Groq OSS 120B} \\ \midrule
E1 C Direct-CoT minus MCU & [-26.25, -26.25] & [5.78, 5.78] & [8.64, 8.64] & [-26.26, -21.26] & NA \\
Mandarin Realis Paradox & [0.00, 0.00] & [75.00, 75.00] & [87.50, 87.50] & [25.00, 25.00] & [0.00, 0.00] \\
English Realis Paradox & [50.00, 50.00] & [0.00, 0.00] & [30.00, 30.00] & [50.00, 50.00] & NA \\
E4 MCU-CF dual minus pure & [15.60, 15.60] & [15.70, 15.70] & [16.04, 16.04] & [6.65, 6.65] & NA \\
\bottomrule\end{tabular}\end{adjustbox}\endgroup\end{center}

Single point values denote equal lower and upper Manski bounds. Rejections (Holm): E2\_\allowbreak{}H1=5/5; E2\_\allowbreak{}H2=1/5; E2\_\allowbreak{}H3=5/5; E2\_\allowbreak{}H4=1/5; E3\_\allowbreak{}H1=4/4; E3\_\allowbreak{}H2=4/4; E3\_\allowbreak{}H4=2/4; E4\_\allowbreak{}H1=1/4; E4\_\allowbreak{}H2=1/4; E4\_\allowbreak{}H3=3/4; E4\_\allowbreak{}H4=3/4.

Discrete NLI-CoT collapse and scalar Direct-CoT collapse dissociate across models evaluated without extended thinking. On DUAL\_\allowbreak{}MOORE, OpenAI E4\_\allowbreak{}H1 is 46.55 pp, whereas Nemotron 4B and 30B yield -19.61 and -20.65 pp. Nevertheless, Nemotron 30B discrete E4\_\allowbreak{}H4 is 75.00 pp (Holm p=9.90973e-05). Gemma E4\_\allowbreak{}H1 is 10.70 pp; its other scalar and discrete contrasts appear alongside the same models in Table 10. The MCU-CF DUAL\_\allowbreak{}MOORE minus PURE\_\allowbreak{}MOORE point estimates are positive only within the four completed E4 stage="dev" model slices examined here; this observation provides no universal guarantee across models or settings. All Nemotron requests set enable\_\allowbreak{}thinking=False, Gemma uses MINIMAL thinking, and no thinking-toggle ablation was performed. Differences in post-training calibration, concise-CoT commitment priors, or single-chain premature lock-in remain unverified mechanistic hypotheses.

For OpenAI DUAL\_\allowbreak{}MOORE, both first\_\allowbreak{}H and mean\_\allowbreak{}H expand from single-output protocols to MCV (Appendix A4c), with MCV-CF draft\_\allowbreak{}copy\_\allowbreak{}rate=0.000; MCV-Direct is 0.050. Thus the zero-copy statement applies to MCV-CF, not all DUAL\_\allowbreak{}MOORE arms. These are top-$K$ truncated-and-renormalized Shannon entropies in bits (OpenAI K=20; Nemotron K=10); absolute cross-K values cannot identify differences in latent uncertainty. Complete-pair diagnostics require status=ok and aligned target logprobs in both arms and use ordinary 95\% percentile bootstrap CIs (B=5,000) with unadjusted two-sided p-values, separate from confirmatory Manski/Holm inference. OpenAI English E1/E3/E4 cells are fully aligned, whereas Mandarin E2 CJK byte-token span alignment yields only partial metric coverage despite semantic 1,000/1,000 ok; Table A4c reports exact per-cell coverage and NA for zero complete pairs. Entropy conclusions are restricted to the fully aligned English cells.

MCV-CF minus Direct-CoT first\_\allowbreak{}H: n=20, difference 1.447 bits, 95\% CI [1.129, 1.725], unadjusted p=0.0004. MCV-CF minus Direct-CoT mean\_\allowbreak{}H: n=20, difference 0.877 bits, 95\% CI [0.598, 1.131], unadjusted p=0.0004. MCV-Direct minus Direct first\_\allowbreak{}H: n=20, difference 1.450 bits, 95\% CI [1.078, 1.830], unadjusted p=0.0004. MCV-Direct minus Direct mean\_\allowbreak{}H: n=20, difference 0.759 bits, 95\% CI [0.455, 1.050], unadjusted p=0.0004.

\begin{figure}[htbp]\centering
\includegraphics[width=0.98\textwidth]{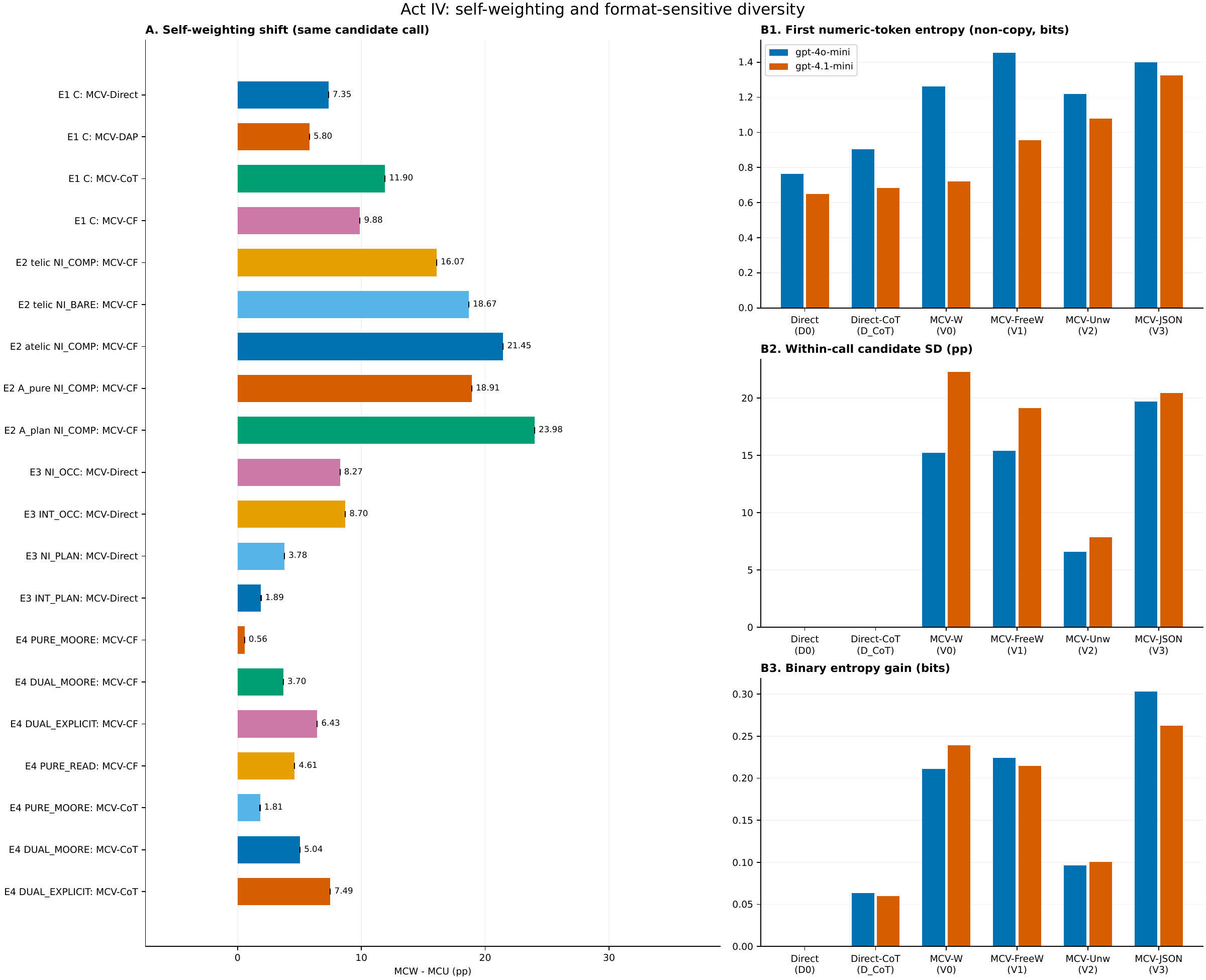}
\caption{Act IV --- Within-call self-weighting shift (MCW minus MCU) across E1--E4 and the 1,800-call OpenAI 6-variant format and token-entropy ablation.}
\end{figure}

\section{Discussion, Limitations, and Conclusion}

Our four-act investigation demonstrates that apparent reasoning failures in aspectual entailment and Moorean pragmatics are systematically governed by the interaction between single-output bottlenecks and evaluation-standard ambiguity. Evaluating discrete three-valued labels (\texttt{NLI-\allowbreak{}Direct} vs. \texttt{NLI-\allowbreak{}CoT}) alongside single-candidate (\texttt{Direct}, \texttt{Direct-\allowbreak{}CoT}) and multi-candidate (\texttt{MCU}, \texttt{MCW}) continuous readouts exposes structural dissociations---most notably the \textbf{Interruption-Anchored Realis Paradox} in Mandarin and English and the \textbf{Directional Reversal} on implicit dual-criteria Moorean assertions---that remain invisible under single-metric benchmarks.

Several methodological limitations bound our conclusions. First, all $16{,}640$ sealed formal inference units in \texttt{E1}--\texttt{E4} were collected on a single local model (\texttt{Qwen3-\allowbreak{}8B-\allowbreak{}Q8\_\allowbreak{}0} at \texttt{temperature=\allowbreak{}0.0}); formal S7 includes 160 paired two-item clusters in E1 and 80 four-slot clusters each in E2--E4. The P16.7 extension comprises 4240 planned/collected units (4222 valid, 18 format-missing, 0 transport-missing) across five external models at 20 clusters per experiment, stage="dev", with less than 2\% missing in each completed slice; OpenAI gpt-4o-mini at top\_\allowbreak{}logprobs=20 has 1,000/1,000 valid units and zero missing. These dev slices broaden model coverage without becoming new formal S7 evidence. Groq slice\_\allowbreak{}e1\_\allowbreak{}e3 and slice\_\allowbreak{}e4 remain incomplete at the 200k TPD free-tier quota boundary; other frozen roster entries remain uncollected. Second, \texttt{E1} operates under the disclosed Option III calibration exception (\texttt{82/\allowbreak{}88}), \texttt{E2}'s atelic subgroup has a modest sample size ($n=32$), and unguided \texttt{MCV-\allowbreak{}Direct} in \texttt{E4} suffers inline-text formatting failures (\texttt{15} missing units in \texttt{E4\_\allowbreak{}H2}) when no reasoning or counterfactual scaffold precedes the five-line numeric block. Third, per \texttt{PLAN-\allowbreak{}E1.md} line 15, our confirmatory contrasts establish differences among observable prompt-and-readout protocols; while consistent with single-chain premature lock-in, they do not isolate internal neural representations or prove that candidate weights reflect calibrated subjective posteriors.

In conclusion, Multi-Candidate Verbalization (\texttt{MCV}) provides a reproducible, zero-training protocol for probing aspectual and pragmatic ambiguity under deterministic decoding. Separating prompt interventions from \texttt{MCU} and \texttt{MCW} readouts allows researchers to diagnose when single-chain protocols exhibit observable standard-selection collapse consistent with the premature lock-in hypothesis.

\appendix
\section{Subclass Profiles, Companion Intervals, and Extension Design}

This appendix reports the complete exploratory subclass profiles for Experiment 2 (Table A1) and Experiment 3 (Table A2, including paired linkage to Experiment 1), alongside the complete 25-metric OpenAI format and token-entropy ablation matrix (Table A3) and the P16.7 protocol, cross-model results and logprob diagnostics (Tables A4a--A4c). All values are computed directly from the twenty-one sealed repository JSON artifacts verified by \texttt{verify\_\allowbreak{}paper\_\allowbreak{}numbers.py}, and the exhaustive machine-indexed claim manifest (all source-bound items) is archived in \texttt{claims\_\allowbreak{}manifest.json} and \texttt{STORY\_\allowbreak{}IMPERFECTIVE.md}.

\subsection{Experiment 2 Six-Subclass Aspectual Summary}

\textbf{Table A1: Experiment 2 Mandarin Six-Subclass Profiles (\texttt{Direct} Means and Paired Contrasts).}

\begin{center}\begingroup\small\setlength{\tabcolsep}{4.5pt}\renewcommand{\arraystretch}{1.12}\begin{adjustbox}{max width=\textwidth}\begin{tabular}{lrrrrrrrr}\toprule
\textbf{Subclass} & \textbf{Clusters (n)} & \textbf{NI\_\allowbreak{}COMP (Direct)} & \textbf{NI\_\allowbreak{}BARE (Direct)} & \textbf{INT\_\allowbreak{}COMP (Direct)} & \textbf{INT\_\allowbreak{}BARE (Direct)} & \textbf{d1: NI\_\allowbreak{}BARE - NI\_\allowbreak{}COMP} & \textbf{d2: Direct - MCU (NI\_\allowbreak{}COMP)} & \textbf{d3: INT\_\allowbreak{}BARE - INT\_\allowbreak{}COMP} \\ \midrule
T\_\allowbreak{}build & 21 & [68.57, 68.57] & [97.99, 97.99] & [2.77, 2.77] & [87.13, 87.13] & [29.42, 29.42] & [13.76, 13.76] & [84.37, 84.37] \\
T\_\allowbreak{}change & 14 & [89.14, 89.14] & [99.51, 99.51] & [7.57, 7.57] & [82.41, 82.41] & [10.37, 10.37] & [28.43, 35.57] & [74.84, 74.84] \\
T\_\allowbreak{}consume & 8 & [83.12, 83.12] & [99.59, 99.59] & [0.00, 0.00] & [83.72, 83.72] & [16.46, 16.46] & [22.18, 22.18] & [83.72, 83.72] \\
T\_\allowbreak{}motion & 5 & [69.90, 69.90] & [97.90, 97.90] & [0.00, 0.00] & [82.98, 82.98] & [28.00, 28.00] & [24.30, 24.30] & [82.98, 82.98] \\
A\_\allowbreak{}pure & 16 & [77.81, 77.81] & [97.58, 97.58] & [15.31, 15.31] & [99.78, 99.78] & [19.77, 19.77] & [23.25, 23.25] & [84.47, 84.47] \\
A\_\allowbreak{}plan & 16 & [81.57, 81.57] & [92.36, 92.36] & [7.50, 7.50] & [99.82, 99.82] & [10.79, 10.79] & [28.66, 28.66] & [92.32, 92.32] \\
\bottomrule\end{tabular}\end{adjustbox}\endgroup\end{center}

\subsection{Experiment 3 Macro-Subclass Breakdown and Experiment 1 Paired Linkage}

\textbf{Table A2: Experiment 3 Macro-Subclass Contrasts (\texttt{A\_\allowbreak{}pure} $n=64$ vs. \texttt{A\_\allowbreak{}plan} $n=16$) and Paired Linkage to Experiment 1 (\texttt{D\_\allowbreak{}orig}, \texttt{B\_\allowbreak{}orig}).}

\begin{center}\begingroup\small\setlength{\tabcolsep}{4.5pt}\renewcommand{\arraystretch}{1.12}\begin{adjustbox}{max width=\textwidth}\begin{tabular}{lrrrr}\toprule
\textbf{Subgroup / Linkage Slice} & \textbf{Quantity / Contrast} & \textbf{Lower Bound (pp)} & \textbf{Upper Bound (pp)} & \textbf{Sample Size (n)} \\ \midrule
E3 A\_\allowbreak{}pure & E3\_\allowbreak{}H1 (bessel\_\allowbreak{}t pp) & 66.28 & 66.28 & 64 \\
E3 A\_\allowbreak{}pure & E3\_\allowbreak{}H2 (bessel\_\allowbreak{}t pp) & 85.81 & 85.81 & 64 \\
E3 A\_\allowbreak{}pure & E3\_\allowbreak{}H3 (bessel\_\allowbreak{}t pp) & 0.67 & 0.67 & 64 \\
E3 A\_\allowbreak{}pure & E3\_\allowbreak{}H4 (exact\_\allowbreak{}tango proportion) & 0.02 & 0.02 & 64 \\
E3 A\_\allowbreak{}plan & E3\_\allowbreak{}H1 (bessel\_\allowbreak{}t pp) & 41.38 & 41.38 & 16 \\
E3 A\_\allowbreak{}plan & E3\_\allowbreak{}H2 (bessel\_\allowbreak{}t pp) & 66.06 & 66.06 & 16 \\
E3 A\_\allowbreak{}plan & E3\_\allowbreak{}H3 (bessel\_\allowbreak{}t pp) & 2.62 & 2.62 & 16 \\
E3 A\_\allowbreak{}plan & E3\_\allowbreak{}H4 (exact\_\allowbreak{}tango proportion) & 0.06 & 0.06 & 16 \\
E3-E1 Linkage (Direct) & NI\_\allowbreak{}OCC\_\allowbreak{}minus\_\allowbreak{}D\_\allowbreak{}orig & -0.88 & -0.88 & 80 \\
E3-E1 Linkage (Direct) & NI\_\allowbreak{}PLAN\_\allowbreak{}minus\_\allowbreak{}D\_\allowbreak{}orig & -62.18 & -62.18 & 80 \\
E3-E1 Linkage (Direct) & INT\_\allowbreak{}OCC\_\allowbreak{}minus\_\allowbreak{}B\_\allowbreak{}orig & -1.68 & -1.68 & 80 \\
E3-E1 Linkage (Direct) & INT\_\allowbreak{}PLAN\_\allowbreak{}minus\_\allowbreak{}B\_\allowbreak{}orig & -83.54 & -83.54 & 80 \\
E3-E1 Linkage (Direct-CoT) & NI\_\allowbreak{}OCC\_\allowbreak{}minus\_\allowbreak{}D\_\allowbreak{}orig & 6.01 & 6.01 & 80 \\
E3-E1 Linkage (Direct-CoT) & NI\_\allowbreak{}PLAN\_\allowbreak{}minus\_\allowbreak{}D\_\allowbreak{}orig & -45.22 & -45.22 & 80 \\
E3-E1 Linkage (Direct-CoT) & INT\_\allowbreak{}OCC\_\allowbreak{}minus\_\allowbreak{}B\_\allowbreak{}orig & 1.02 & 1.02 & 80 \\
E3-E1 Linkage (Direct-CoT) & INT\_\allowbreak{}PLAN\_\allowbreak{}minus\_\allowbreak{}B\_\allowbreak{}orig & -79.39 & -79.39 & 80 \\
E3-E1 Linkage (MCU) & NI\_\allowbreak{}OCC\_\allowbreak{}minus\_\allowbreak{}D\_\allowbreak{}orig & -5.52 & -5.52 & 80 \\
E3-E1 Linkage (MCU) & NI\_\allowbreak{}PLAN\_\allowbreak{}minus\_\allowbreak{}D\_\allowbreak{}orig & -59.48 & -59.48 & 80 \\
E3-E1 Linkage (MCU) & INT\_\allowbreak{}OCC\_\allowbreak{}minus\_\allowbreak{}B\_\allowbreak{}orig & -1.77 & -1.77 & 80 \\
E3-E1 Linkage (MCU) & INT\_\allowbreak{}PLAN\_\allowbreak{}minus\_\allowbreak{}B\_\allowbreak{}orig & -56.40 & -56.40 & 80 \\
E3-E1 Linkage (MCU-CoT) & NI\_\allowbreak{}OCC\_\allowbreak{}minus\_\allowbreak{}D\_\allowbreak{}orig & 3.73 & 4.98 & 80 \\
E3-E1 Linkage (MCU-CoT) & NI\_\allowbreak{}PLAN\_\allowbreak{}minus\_\allowbreak{}D\_\allowbreak{}orig & -43.93 & -43.93 & 80 \\
E3-E1 Linkage (MCU-CoT) & INT\_\allowbreak{}OCC\_\allowbreak{}minus\_\allowbreak{}B\_\allowbreak{}orig & 6.17 & 8.67 & 80 \\
E3-E1 Linkage (MCU-CoT) & INT\_\allowbreak{}PLAN\_\allowbreak{}minus\_\allowbreak{}B\_\allowbreak{}orig & -53.08 & -51.83 & 80 \\
\bottomrule\end{tabular}\end{adjustbox}\endgroup\end{center}

\subsection{Complete OpenAI 1,800-Call Format and Token-Entropy Ablation Tables}

\textbf{Table A3a: Calibration, Discrimination, and Readout Contrasts Across Six Format Variants.}

\begin{center}\begingroup\small\setlength{\tabcolsep}{4.5pt}\renewcommand{\arraystretch}{1.12}\begin{adjustbox}{max width=\textwidth}\begin{tabular}{lllllll}\toprule
\textbf{Model} & \textbf{Variant} & \textbf{Complete Pairs} & \textbf{H2 Hedging} & \textbf{H4a Interaction} & \textbf{AUC (Stated / Reported / Overall / Delta)} & \textbf{H4b vs. Direct (MCW / MCU / Median / Top-1)} \\ \midrule
gpt-4o-mini & Direct (D0) & 30 & -5.33 & 9.00 & 0.92 / 0.89 / 0.91 / 0.03 & 0.00 / 0.00 / 0.00 / 0.00 \\
gpt-4o-mini & Direct-CoT (D\_\allowbreak{}CoT) & 30 & -1.67 & 11.67 & 0.94 / 0.89 / 0.92 / 0.05 & -2.37 / -2.37 / -2.37 / -2.37 \\
gpt-4o-mini & MCV-Weighted (V0) & 30 & 0.92 & 8.97 & 0.90 / 0.81 / 0.85 / 0.09 & -8.67 / -17.62 / -19.33 / 3.23 \\
gpt-4o-mini & MCV-FreeWeight (V1) & 30 & -0.93 & 8.36 & 0.90 / 0.85 / 0.88 / 0.04 & -5.58 / -15.57 / -17.01 / 4.77 \\
gpt-4o-mini & MCV-UnweightedPrompt (V2) & 30 & -0.96 & 9.43 & 0.92 / 0.90 / 0.91 / 0.02 & -5.23 / -5.23 / -5.27 / -4.77 \\
gpt-4o-mini & MCV-JSON (V3) & 30 & -1.37 & 5.16 & 0.82 / 0.77 / 0.79 / 0.04 & -3.20 / -12.33 / -13.80 / 8.70 \\
gpt-4.1-mini & Direct (D0) & 30 & -4.72 & 13.76 & 0.91 / 0.82 / 0.86 / 0.09 & 0.00 / 0.00 / 0.00 / 0.00 \\
gpt-4.1-mini & Direct-CoT (D\_\allowbreak{}CoT) & 30 & -0.50 & 15.00 & 0.94 / 0.90 / 0.92 / 0.04 & -1.83 / -1.83 / -1.83 / -1.83 \\
gpt-4.1-mini & MCV-Weighted (V0) & 30 & -0.52 & 12.77 & 0.92 / 0.79 / 0.86 / 0.12 & -8.79 / -10.77 / -12.63 / -3.35 \\
gpt-4.1-mini & MCV-FreeWeight (V1) & 30 & -1.18 & 1.50 & 0.87 / 0.89 / 0.88 / -0.02 & -6.59 / -12.01 / -13.68 / 2.50 \\
gpt-4.1-mini & MCV-UnweightedPrompt (V2) & 30 & -0.30 & 11.76 & 0.93 / 0.89 / 0.91 / 0.03 & -5.05 / -5.05 / -5.02 / -6.63 \\
gpt-4.1-mini & MCV-JSON (V3) & 30 & -1.75 & 9.71 & 0.89 / 0.85 / 0.88 / 0.04 & -0.77 / -2.82 / -3.43 / 3.08 \\
\bottomrule\end{tabular}\end{adjustbox}\endgroup\end{center}

\textbf{Table A3b: Endpoint Mass, Weight Formatting Diagnostics, and Token-Level Shannon Entropies.}

\begin{center}\begingroup\small\setlength{\tabcolsep}{4.5pt}\renewcommand{\arraystretch}{1.12}\begin{adjustbox}{max width=\textwidth}\begin{tabular}{lllllll}\toprule
\textbf{Model} & \textbf{Variant} & \textbf{Endpoint Mass (Prop.)} & \textbf{Weight-Sum In-Tol (Prop.)} & \textbf{Draft-Then-Copy (Prop.)} & \textbf{Mean Tokens} & \textbf{Token Entropy: First-Num / Mean-Num / NonCopy-Mean-Num / All-Token (bits)} \\ \midrule
gpt-4o-mini & Direct (D0) & 0.13 & 1.00 & 0.00 & 93.32 & 0.678 / 0.678 / 0.765 / 0.742 \\
gpt-4o-mini & Direct-CoT (D\_\allowbreak{}CoT) & 0.09 & 1.00 & 0.00 & 94.55 & 0.879 / 0.879 / 0.903 / 0.899 \\
gpt-4o-mini & MCV-Weighted (V0) & 0.06 & 1.00 & 0.09 & 175.09 & 1.151 / 0.961 / 1.052 / 0.558 \\
gpt-4o-mini & MCV-FreeWeight (V1) & 0.04 & 0.89 & 0.00 & 83.77 & 1.454 / 1.247 / 1.247 / 0.189 \\
gpt-4o-mini & MCV-UnweightedPrompt (V2) & 0.09 & 1.00 & 0.00 & 106.80 & 1.219 / 1.029 / 1.029 / 0.629 \\
gpt-4o-mini & MCV-JSON (V3) & 0.02 & 0.99 & 0.00 & 61.02 & 1.400 / 0.895 / 0.895 / 0.118 \\
gpt-4.1-mini & Direct (D0) & 0.18 & 1.00 & 0.00 & 137.93 & 0.588 / 0.588 / 0.648 / 0.810 \\
gpt-4.1-mini & Direct-CoT (D\_\allowbreak{}CoT) & 0.11 & 1.00 & 0.00 & 101.36 & 0.679 / 0.679 / 0.683 / 0.759 \\
gpt-4.1-mini & MCV-Weighted (V0) & 0.18 & 1.00 & 0.76 & 368.90 & 0.120 / 0.069 / 0.393 / 0.698 \\
gpt-4.1-mini & MCV-FreeWeight (V1) & 0.16 & 0.81 & 0.75 & 359.78 & 0.172 / 0.153 / 0.851 / 0.705 \\
gpt-4.1-mini & MCV-UnweightedPrompt (V2) & 0.11 & 1.00 & 0.26 & 237.02 & 0.791 / 0.614 / 0.838 / 0.770 \\
gpt-4.1-mini & MCV-JSON (V3) & 0.08 & 1.00 & 0.00 & 80.00 & 1.325 / 0.768 / 0.768 / 0.077 \\
\bottomrule\end{tabular}\end{adjustbox}\endgroup\end{center}

\subsection{P16.7 Protocol, Cross-Model Bounds, and Token Diagnostics}

\textbf{Table A4a: Frozen P16.7/OpenAI addendum protocol and probe gate.}

\begin{center}\begingroup\small\setlength{\tabcolsep}{4.5pt}\renewcommand{\arraystretch}{1.12}\begin{adjustbox}{max width=\textwidth}\begin{tabular}{lr}\toprule
\textbf{Source / parameter} & \textbf{Value} \\ \midrule
API / stage & dev \\
API\_\allowbreak{}ADD / slices/all/count & 1000 \\
API\_\allowbreak{}ADD / statistics/B & 5000 \\
openai:gpt-4o-mini / top\_\allowbreak{}logprobs & 20 \\
openai:gpt-4o-mini / max\_\allowbreak{}tokens & 1024 \\
nv:nemotron-3.5-lightning-30b / top\_\allowbreak{}logprobs & 10 \\
OpenAI probe / n\_\allowbreak{}units & 12 \\
OpenAI probe / n\_\allowbreak{}target\_\allowbreak{}positions\_\allowbreak{}checked & 28 \\
OpenAI probe / min\_\allowbreak{}target\_\allowbreak{}top\_\allowbreak{}logprobs & 20 \\
OpenAI probe / max\_\allowbreak{}target\_\allowbreak{}top\_\allowbreak{}logprobs & 20 \\
\bottomrule\end{tabular}\end{adjustbox}\endgroup\end{center}

Deterministic temperature=0.0 and OpenAI top\_\allowbreak{}p=1.0 are locked in the frozen collector source (hash sealed by API\_\allowbreak{}ADD); all Nemotron requests set enable\_\allowbreak{}thinking=False; Gemma uses MINIMAL thinking. No thinking-toggle ablation was performed.

\textbf{Table A4b: Complete primary slot bounds; simultaneous 98.75\% CI; Holm p.}

\begin{center}\begingroup\small\setlength{\tabcolsep}{4.5pt}\renewcommand{\arraystretch}{1.12}\begin{adjustbox}{max width=\textwidth}\begin{tabular}{lrrrrr}\toprule
\textbf{E2 slot: bounds; CI; p} & \textbf{OpenAI 4o-mini} & \textbf{Gemma 31B} & \textbf{Nemotron 4B} & \textbf{Nemotron 30B} & \textbf{Groq OSS 120B} \\ \midrule
E2\_\allowbreak{}H1 & 40.42; [25.42, 54.17]; 0.0008 & 24.50; [9.25, 44.50]; 0.0008 & 32.75; [15.42, 51.67]; 0.0008 & 27.52; [1.84, 55.32]; 0.0008 & 45.32; [32.48, 58.39]; 0.0008 \\
E2\_\allowbreak{}H2 & 15.00; [5.75, 26.75]; 0.0008 & 5.27; [-7.92, 16.10]; 0.2827 & -1.08; [-22.75, 21.83]; 1.0000 & [6.48, 14.81]; [-16.72, 43.62]; 0.2374 & -1.50; [-12.83, 13.00]; 1.0000 \\
E2\_\allowbreak{}H3 & 71.88; [38.75, 95.62]; 0.0008 & 98.00; [96.12, 99.88]; 0.0008 & 86.88; [51.25, 100.00]; 0.0008 & 59.31; [18.75, 96.75]; 0.0012 & 94.75; [89.25, 99.25]; 0.0008 \\
E2\_\allowbreak{}H4 & 0.00; [-48.30, 48.30]; 1.0000 & 25.00; [-31.53, 73.36]; 0.2827 & 12.50; [-46.46, 64.46]; 0.5516 & 37.50; [-26.26, 82.80]; 0.1122 & 100.00; [5.91, 100.00]; 0.0078 \\
\bottomrule\end{tabular}\end{adjustbox}\endgroup\end{center}

\begin{center}\begingroup\small\setlength{\tabcolsep}{4.5pt}\renewcommand{\arraystretch}{1.12}\begin{adjustbox}{max width=\textwidth}\begin{tabular}{lrrrrr}\toprule
\textbf{E3 slot: bounds; CI; p} & \textbf{OpenAI 4o-mini} & \textbf{Gemma 31B} & \textbf{Nemotron 4B} & \textbf{Nemotron 30B} & \textbf{Groq OSS 120B} \\ \midrule
E3\_\allowbreak{}H1 & 56.45; [44.70, 67.50]; 0.0008 & 38.70; [21.25, 58.85]; 0.0006 & 44.95; [30.55, 59.05]; 0.0008 & 55.45; [35.15, 74.45]; 0.0008 & uncollected \\
E3\_\allowbreak{}H2 & 78.50; [69.75, 83.50]; 0.0008 & 85.30; [68.95, 96.50]; 0.0006 & 75.40; [59.65, 87.45]; 0.0008 & 80.25; [62.50, 91.75]; 0.0008 & uncollected \\
E3\_\allowbreak{}H3 & [-100.00, 100.00]; [-100.00, 100.00]; 1.0000 & [-100.00, 100.00]; [-100.00, 100.00]; 1.0000 & [-100.00, 100.00]; [-100.00, 100.00]; 1.0000 & [-100.00, 100.00]; [-100.00, 100.00]; 1.0000 & uncollected \\
E3\_\allowbreak{}H4 & 5.00; [-21.70, 32.29]; 0.5385 & 100.00; [55.06, 100.00]; 3.81470e-06 & 5.00; [-21.70, 32.29]; 0.5385 & 50.00; [11.64, 77.67]; 0.0009 & uncollected \\
\bottomrule\end{tabular}\end{adjustbox}\endgroup\end{center}

\begin{center}\begingroup\small\setlength{\tabcolsep}{4.5pt}\renewcommand{\arraystretch}{1.12}\begin{adjustbox}{max width=\textwidth}\begin{tabular}{lrrrrr}\toprule
\textbf{E4 slot: bounds; CI; p} & \textbf{OpenAI 4o-mini} & \textbf{Gemma 31B} & \textbf{Nemotron 4B} & \textbf{Nemotron 30B} & \textbf{Groq OSS 120B} \\ \midrule
E4\_\allowbreak{}H1 & 46.55; [24.90, 64.60]; 0.0008 & 10.70; [-11.05, 32.45]; 0.1120 & -19.61; [-23.82, -15.61]; 1.0000 & -20.65; [-35.70, 0.10]; 1.0000 & uncollected \\
E4\_\allowbreak{}H2 & 29.20; [19.80, 37.40]; 0.0008 & 18.38; [-8.16, 40.75]; 0.0612 & -8.47; [-22.33, 5.42]; 1.0000 & [-0.80, 14.20]; [-12.30, 30.90]; 1.0000 & uncollected \\
E4\_\allowbreak{}H3 & 72.25; [50.00, 90.75]; 0.0008 & 37.75; [14.75, 61.75]; 0.0006 & 19.00; [0.20, 42.70]; 0.0128 & [0.00, 5.00]; [-19.50, 19.50]; 1.0000 & uncollected \\
E4\_\allowbreak{}H4 & 55.00; [19.14, 82.37]; 0.0008 & 90.00; [49.93, 99.41]; 7.57755e-06 & [10.00, 15.00]; [-20.42, 47.40]; 0.3307 & 75.00; [36.72, 93.89]; 9.90973e-05 & uncollected \\
\bottomrule\end{tabular}\end{adjustbox}\endgroup\end{center}

E3\_\allowbreak{}H3 is descriptive and intentionally uncollected: E3 MCV-CoT is omitted from slice\_\allowbreak{}e1\_\allowbreak{}e3. Its [-100.00, 100.00] pp Manski interval is not an observed null result and does not add a confirmatory rejection. Key diagnostics appear in Table 10; complete diagnostic and E3 three-label counts remain source-bound in the story ledger.

Scale audit: 13 continuous diagnostics in frozen analyze\_\allowbreak{}p167.py passed unnormalized multinomial cluster counts (summing to n=8, 12 or 20) to infer\_\allowbreak{}continuous\_\allowbreak{}slot routines. Their bootstrap intervals and spans (ci\_\allowbreak{}9875, ci\_\allowbreak{}95, raw\_\allowbreak{}ci\_\allowbreak{}\textit{, exploratory\_\allowbreak{}ci\_\allowbreak{}9875 and span\_\allowbreak{}}) are on the cluster-sum scale n * pp, not pp. These fields, together with dimensionless p / exploratory\_\allowbreak{}p\_\allowbreak{}raw and resampling/method metadata from the same auxiliary inference package, are conservatively excluded from the verified claim manifest, story ledger and all scientific inference. Only integer counts n / missing\_\allowbreak{}units and verified pp point\_\allowbreak{}lower / point\_\allowbreak{}upper / point\_\allowbreak{}bounds are retained for these diagnostics. The engines E2/E3/E4 primary\_\allowbreak{}family results use internally normalized weights and are unaffected.

\textbf{Table A4c: Logprob coverage and exploratory entropy diagnostics (bits).}

\begin{center}\begingroup\small\setlength{\tabcolsep}{4.5pt}\renewcommand{\arraystretch}{1.12}\begin{adjustbox}{max width=\textwidth}\begin{tabular}{llll}\toprule
\textbf{Label cell / q(True) lower, upper; captured mass; H} & \textbf{OpenAI 4o-mini} & \textbf{Nemotron 4B} & \textbf{Nemotron 30B} \\ \midrule
E3/NI\_\allowbreak{}OCC NLI-Direct & [0.894, 0.894]; 1.000 (20/20); 0.336 (20/20) & [0.247, 0.247]; 1.000 (20/20); NA (0/20) & [0.900, 0.906]; 0.994 (20/20); 0.506 (20/20) \\
E3/NI\_\allowbreak{}PLAN NLI-Direct & [0.000, 0.000]; 1.000 (20/20); 0.000 (20/20) & [0.000, 0.000]; 1.000 (20/20); NA (0/20) & [0.069, 0.076]; 0.994 (20/20); 0.919 (20/20) \\
E3/INT\_\allowbreak{}OCC NLI-Direct & [0.870, 0.870]; 1.000 (20/20); 0.150 (20/20) & [0.423, 0.423]; 0.999 (20/20); NA (0/20) & [0.902, 0.910]; 0.993 (20/20); 0.514 (20/20) \\
E3/INT\_\allowbreak{}PLAN NLI-Direct & [0.000, 0.000]; 1.000 (20/20); 0.141 (20/20) & [0.000, 0.000]; 1.000 (20/20); NA (0/20) & [0.046, 0.057]; 0.989 (20/20); 0.819 (20/20) \\
E2/INT\_\allowbreak{}BARE NLI-CoT & [0.001, 0.001]; 1.000 (4/20); 0.049 (4/20) & [0.500, 0.500]; 1.000 (20/20); NA (0/20) & [NA, NA]; NA (0/20); NA (0/20) \\
E2/INT\_\allowbreak{}COMP NLI-CoT & [0.000, 0.000]; 1.000 (1/20); 0.000 (1/20) & [0.000, 0.000]; 1.000 (20/20); NA (0/20) & [NA, NA]; NA (0/20); NA (0/20) \\
E2/NI\_\allowbreak{}BARE NLI-CoT & [0.000, 0.000]; 1.000 (12/20); 0.000 (12/20) & [0.050, 0.050]; 1.000 (20/20); NA (0/20) & [NA, NA]; NA (0/20); NA (0/20) \\
E2/NI\_\allowbreak{}COMP NLI-CoT & [0.000, 0.000]; 1.000 (12/20); 0.110 (12/20) & [0.000, 0.000]; 1.000 (20/20); NA (0/20) & [NA, NA]; NA (0/20); NA (0/20) \\
E3/INT\_\allowbreak{}OCC NLI-CoT & [0.550, 0.550]; 1.000 (20/20); 0.000 (20/20) & [0.350, 0.350]; 1.000 (20/20); NA (0/20) & [1.000, 1.000]; 1.000 (20/20); 0.000 (20/20) \\
E3/INT\_\allowbreak{}PLAN NLI-CoT & [0.000, 0.000]; 1.000 (20/20); 0.100 (20/20) & [0.000, 0.000]; 1.000 (20/20); NA (0/20) & [0.000, 0.000]; 1.000 (20/20); 0.041 (20/20) \\
E3/NI\_\allowbreak{}OCC NLI-CoT & [0.050, 0.050]; 1.000 (20/20); 0.000 (20/20) & [0.050, 0.050]; 1.000 (20/20); NA (0/20) & [0.500, 0.500]; 1.000 (20/20); 0.000 (20/20) \\
E3/NI\_\allowbreak{}PLAN NLI-CoT & [0.000, 0.000]; 1.000 (20/20); 0.000 (20/20) & [0.000, 0.000]; 1.000 (20/20); NA (0/20) & [0.000, 0.000]; 1.000 (20/20); 0.001 (20/20) \\
E4/DUAL\_\allowbreak{}EXPLICIT NLI-CoT & [0.550, 0.550]; 1.000 (20/20); 0.000 (20/20) & [NA, NA]; NA (0/20); NA (0/20) & [0.950, 0.950]; 1.000 (20/20); 0.001 (20/20) \\
E4/DUAL\_\allowbreak{}MOORE NLI-CoT & [0.000, 0.000]; 1.000 (20/20); 0.000 (20/20) & [NA, NA]; NA (0/20); NA (0/20) & [0.200, 0.200]; 1.000 (20/20); 0.001 (20/20) \\
E4/PURE\_\allowbreak{}MOORE NLI-CoT & [0.000, 0.000]; 1.000 (20/20); 0.000 (20/20) & [NA, NA]; NA (0/20); NA (0/20) & [0.156, 0.156]; 1.000 (20/20); 0.027 (20/20) \\
E4/PURE\_\allowbreak{}READ NLI-CoT & [0.000, 0.000]; 1.000 (20/20); 0.000 (20/20) & [NA, NA]; NA (0/20); NA (0/20) & [0.944, 0.944]; 1.000 (18/20); 0.000 (18/20) \\
\bottomrule\end{tabular}\end{adjustbox}\endgroup\end{center}

q(True) bounds and captured mass share the mass coverage; H has its own metric coverage.

\begin{center}\begingroup\small\setlength{\tabcolsep}{4.5pt}\renewcommand{\arraystretch}{1.12}\begin{adjustbox}{max width=\textwidth}\begin{tabular}{llll}\toprule
\textbf{Numeric cell: first H / mean H; draft-copy} & \textbf{OpenAI 4o-mini} & \textbf{Nemotron 4B} & \textbf{Nemotron 30B} \\ \midrule
E1/C MCV-Direct & 0.823 (20/20) / 0.770 (20/20); 0.000 & NA (0/20) / NA (0/20); NA & 0.745 (19/20) / 0.564 (19/20); 0.211 \\
E1/C Direct-CoT & 1.111 (20/20) / 1.111 (20/20); 0.050 & NA (0/20) / NA (0/20); NA & 0.731 (20/20) / 0.731 (20/20); 0.000 \\
E2/INT\_\allowbreak{}BARE MCV-Direct & 1.485 (3/20) / 1.359 (3/20); 0.000 & NA (0/20) / NA (0/20); NA & 2.808 (11/20) / 1.901 (11/20); 0.000 \\
E2/INT\_\allowbreak{}BARE Direct & 1.008 (3/20) / 1.008 (3/20); 0.500 & NA (0/20) / NA (0/20); NA & NA (0/20) / NA (0/20); 0.250 \\
E2/INT\_\allowbreak{}COMP MCV-Direct & 0.559 (3/20) / 0.679 (3/20); 0.000 & NA (0/20) / NA (0/20); NA & 2.736 (10/20) / 1.782 (10/20); 0.000 \\
E2/INT\_\allowbreak{}COMP Direct & 0.000 (2/20) / 0.000 (2/20); 0.800 & NA (0/20) / NA (0/20); NA & NA (0/20) / NA (0/20); 0.000 \\
E2/NI\_\allowbreak{}BARE MCV-Direct & 0.820 (7/20) / 0.655 (7/20); 0.000 & NA (0/20) / NA (0/20); NA & 2.386 (10/20) / 1.700 (10/20); 0.000 \\
E2/NI\_\allowbreak{}BARE Direct & 0.687 (7/20) / 0.687 (7/20); 0.450 & NA (0/20) / NA (0/20); NA & NA (0/20) / NA (0/20); 0.100 \\
E2/NI\_\allowbreak{}COMP MCV-Direct & 1.771 (7/20) / 1.168 (7/20); 0.050 & NA (0/20) / NA (0/20); NA & 2.677 (9/20) / 1.699 (9/20); 0.000 \\
E2/NI\_\allowbreak{}COMP Direct & 0.949 (5/20) / 0.949 (5/20); 0.400 & NA (0/20) / NA (0/20); NA & NA (0/20) / NA (0/20); 0.150 \\
E3/INT\_\allowbreak{}OCC MCV-Direct & 0.875 (20/20) / 0.935 (20/20); 0.000 & NA (0/20) / NA (0/20); NA & 1.333 (19/20) / 0.775 (19/20); 0.211 \\
E3/INT\_\allowbreak{}OCC Direct & 0.792 (20/20) / 0.792 (20/20); 0.000 & NA (0/20) / NA (0/20); NA & 0.348 (20/20) / 0.348 (20/20); 0.000 \\
E3/INT\_\allowbreak{}PLAN MCV-Direct & 1.199 (20/20) / 1.097 (20/20); 0.050 & NA (0/20) / NA (0/20); NA & 1.038 (20/20) / 0.540 (20/20); 0.350 \\
E3/INT\_\allowbreak{}PLAN Direct & 0.783 (20/20) / 0.783 (20/20); 0.050 & NA (0/20) / NA (0/20); NA & 1.026 (20/20) / 1.026 (20/20); 0.000 \\
E3/NI\_\allowbreak{}OCC MCV-Direct & 0.560 (20/20) / 0.916 (20/20); 0.000 & NA (0/20) / NA (0/20); NA & 0.864 (20/20) / 0.650 (20/20); 0.350 \\
E3/NI\_\allowbreak{}OCC Direct & 0.615 (20/20) / 0.615 (20/20); 0.000 & NA (0/20) / NA (0/20); NA & 0.341 (20/20) / 0.341 (20/20); 0.000 \\
E3/NI\_\allowbreak{}PLAN MCV-Direct & 1.899 (20/20) / 0.801 (20/20); 0.200 & NA (0/20) / NA (0/20); NA & 1.418 (20/20) / 0.765 (20/20); 0.350 \\
E3/NI\_\allowbreak{}PLAN Direct & 1.162 (20/20) / 1.162 (20/20); 0.150 & NA (0/20) / NA (0/20); NA & 1.281 (20/20) / 1.281 (20/20); 0.050 \\
E4/DUAL\_\allowbreak{}EXPLICIT MCV-CF & 1.714 (20/20) / 1.275 (20/20); 0.000 & NA (0/20) / NA (0/20); 0.050 & 1.442 (20/20) / 0.741 (20/20); 0.000 \\
E4/DUAL\_\allowbreak{}EXPLICIT MCV-Direct & 1.749 (20/20) / 1.082 (20/20); 0.050 & NA (0/20) / NA (0/20); 0.263 & 1.611 (20/20) / 0.650 (20/20); 0.000 \\
E4/DUAL\_\allowbreak{}EXPLICIT Direct-CoT & 0.670 (20/20) / 0.670 (20/20); 0.000 & NA (0/20) / NA (0/20); 0.000 & 0.854 (20/20) / 0.854 (20/20); 0.000 \\
E4/DUAL\_\allowbreak{}EXPLICIT Direct & 0.872 (20/20) / 0.872 (20/20); 0.000 & NA (0/20) / NA (0/20); 0.450 & 1.403 (19/20) / 1.403 (19/20); 0.053 \\
E4/DUAL\_\allowbreak{}MOORE MCV-CF & 1.788 (20/20) / 1.219 (20/20); 0.000 & NA (0/20) / NA (0/20); 0.450 & 1.619 (20/20) / 0.787 (20/20); 0.000 \\
E4/DUAL\_\allowbreak{}MOORE MCV-Direct & 1.872 (20/20) / 1.181 (20/20); 0.050 & NA (0/20) / NA (0/20); 0.550 & 1.715 (17/20) / 0.654 (17/20); 0.235 \\
E4/DUAL\_\allowbreak{}MOORE Direct-CoT & 0.341 (20/20) / 0.341 (20/20); 0.000 & NA (0/20) / NA (0/20); 0.000 & 0.887 (20/20) / 0.887 (20/20); 0.000 \\
E4/DUAL\_\allowbreak{}MOORE Direct & 0.422 (20/20) / 0.422 (20/20); 0.050 & NA (0/20) / NA (0/20); 0.100 & 2.523 (20/20) / 2.523 (20/20); 0.000 \\
E4/PURE\_\allowbreak{}MOORE MCV-CF & 1.801 (20/20) / 1.271 (20/20); 0.000 & NA (0/20) / NA (0/20); 0.150 & 1.600 (20/20) / 0.905 (20/20); 0.000 \\
E4/PURE\_\allowbreak{}MOORE MCV-Direct & 1.001 (20/20) / 1.093 (20/20); 0.000 & NA (0/20) / NA (0/20); 0.400 & 1.860 (20/20) / 0.859 (20/20); 0.300 \\
E4/PURE\_\allowbreak{}MOORE Direct-CoT & 0.001 (20/20) / 0.001 (20/20); 0.050 & NA (0/20) / NA (0/20); 0.000 & 1.124 (20/20) / 1.124 (20/20); 0.000 \\
E4/PURE\_\allowbreak{}MOORE Direct & 0.118 (20/20) / 0.118 (20/20); 0.250 & NA (0/20) / NA (0/20); 0.250 & 2.676 (20/20) / 2.676 (20/20); 0.000 \\
E4/PURE\_\allowbreak{}READ MCV-CF & 1.648 (20/20) / 1.265 (20/20); 0.000 & NA (0/20) / NA (0/20); 0.050 & 1.134 (19/20) / 0.838 (19/20); 0.000 \\
E4/PURE\_\allowbreak{}READ MCV-Direct & 1.021 (20/20) / 1.197 (20/20); 0.000 & NA (0/20) / NA (0/20); 0.250 & 2.135 (20/20) / 1.169 (20/20); 0.000 \\
E4/PURE\_\allowbreak{}READ Direct-CoT & 0.053 (20/20) / 0.053 (20/20); 0.000 & NA (0/20) / NA (0/20); 0.000 & 1.004 (20/20) / 1.004 (20/20); 0.000 \\
E4/PURE\_\allowbreak{}READ Direct & 0.192 (20/20) / 0.192 (20/20); 0.100 & NA (0/20) / NA (0/20); 0.150 & 2.207 (20/20) / 2.207 (20/20); 0.000 \\
\bottomrule\end{tabular}\end{adjustbox}\endgroup\end{center}

\begin{center}\begingroup\small\setlength{\tabcolsep}{4.5pt}\renewcommand{\arraystretch}{1.12}\begin{adjustbox}{max width=\textwidth}\begin{tabular}{llll}\toprule
\textbf{Complete-pair contrast / metric: n; estimate [95\% CI]; raw p} & \textbf{OpenAI 4o-mini} & \textbf{Nemotron 4B} & \textbf{Nemotron 30B} \\ \midrule
E3\_\allowbreak{}NI\_\allowbreak{}OCC MCV-Direct minus Direct\_\allowbreak{}first\_\allowbreak{}H & 20; -0.055 [-0.195, 0.079]; 0.4471 & 0; NA NA; NA & 20; 0.523 [0.195, 0.865]; 0.0020 \\
E3\_\allowbreak{}NI\_\allowbreak{}OCC MCV-Direct minus Direct\_\allowbreak{}mean\_\allowbreak{}H & 20; 0.301 [0.091, 0.492]; 0.0068 & 0; NA NA; NA & 20; 0.309 [0.047, 0.575]; 0.0224 \\
E3\_\allowbreak{}INT\_\allowbreak{}OCC MCV-Direct minus Direct\_\allowbreak{}first\_\allowbreak{}H & 20; 0.083 [-0.061, 0.225]; 0.2543 & 0; NA NA; NA & 19; 0.970 [0.581, 1.369]; 0.0004 \\
E3\_\allowbreak{}INT\_\allowbreak{}OCC MCV-Direct minus Direct\_\allowbreak{}mean\_\allowbreak{}H & 20; 0.143 [-0.028, 0.305]; 0.1028 & 0; NA NA; NA & 19; 0.411 [0.184, 0.647]; 0.0008 \\
E2\_\allowbreak{}TELIC\_\allowbreak{}NI\_\allowbreak{}COMP MCV-Direct minus Direct\_\allowbreak{}first\_\allowbreak{}H & 1; 0.764 [0.764, 0.764]; 0.0004 & 0; NA NA; NA & 0; NA NA; NA \\
E2\_\allowbreak{}TELIC\_\allowbreak{}NI\_\allowbreak{}COMP MCV-Direct minus Direct\_\allowbreak{}mean\_\allowbreak{}H & 1; -0.111 [-0.111, -0.111]; 0.0004 & 0; NA NA; NA & 0; NA NA; NA \\
E2\_\allowbreak{}ATELIC\_\allowbreak{}INT\_\allowbreak{}COMP MCV-Direct minus Direct\_\allowbreak{}first\_\allowbreak{}H & 0; NA NA; NA & 0; NA NA; NA & 0; NA NA; NA \\
E2\_\allowbreak{}ATELIC\_\allowbreak{}INT\_\allowbreak{}COMP MCV-Direct minus Direct\_\allowbreak{}mean\_\allowbreak{}H & 0; NA NA; NA & 0; NA NA; NA & 0; NA NA; NA \\
E4\_\allowbreak{}DUAL\_\allowbreak{}MOORE MCV-Direct minus Direct\_\allowbreak{}first\_\allowbreak{}H & 20; 1.450 [1.078, 1.830]; 0.0004 & 0; NA NA; NA & 17; -0.808 [-1.356, -0.326]; 0.0004 \\
E4\_\allowbreak{}DUAL\_\allowbreak{}MOORE MCV-Direct minus Direct\_\allowbreak{}mean\_\allowbreak{}H & 20; 0.759 [0.455, 1.050]; 0.0004 & 0; NA NA; NA & 17; -1.869 [-2.123, -1.640]; 0.0004 \\
E4\_\allowbreak{}DUAL\_\allowbreak{}MOORE MCV-CF minus Direct-CoT\_\allowbreak{}first\_\allowbreak{}H & 20; 1.447 [1.129, 1.725]; 0.0004 & 0; NA NA; NA & 20; 0.732 [0.498, 0.967]; 0.0004 \\
E4\_\allowbreak{}DUAL\_\allowbreak{}MOORE MCV-CF minus Direct-CoT\_\allowbreak{}mean\_\allowbreak{}H & 20; 0.877 [0.598, 1.131]; 0.0004 & 0; NA NA; NA & 20; -0.100 [-0.297, 0.103]; 0.3347 \\
E1\_\allowbreak{}C MCV-Direct minus Direct-CoT\_\allowbreak{}first\_\allowbreak{}H & 20; -0.288 [-0.619, 0.077]; 0.1204 & 0; NA NA; NA & 19; -0.019 [-0.475, 0.464]; 0.9406 \\
E1\_\allowbreak{}C MCV-Direct minus Direct-CoT\_\allowbreak{}mean\_\allowbreak{}H & 20; -0.341 [-0.571, -0.098]; 0.0072 & 0; NA NA; NA & 19; -0.200 [-0.607, 0.205]; 0.3351 \\
\bottomrule\end{tabular}\end{adjustbox}\endgroup\end{center}

All token entropies are top-$K$ truncated-and-renormalized Shannon entropies in bits: OpenAI K=20, Nemotron K=10. Absolute cross-K differences must not be interpreted as differences in latent model uncertainty; only within-model fixed-K contrasts are directly comparable. Each metric gives its own n\_\allowbreak{}available/n\_\allowbreak{}planned; NA (0/20) means unavailable, never zero entropy. Nemotron 4B E1/E3 and E2 use the legacy runner, whereas E4 uses the addendum runner: q(True) can have 20/20 coverage while label entropy has 0/20. OpenAI English E1/E3/E4 cells have full alignment; Mandarin E2 uses strict UTF-8 CJK byte-token span alignment and has partial coverage (L/CoT 1..12/20; numeric 2..7/20). Semantic ok counts do not substitute for metric coverage; no zero-padding is used. E2\_\allowbreak{}ATELIC\_\allowbreak{}INT\_\allowbreak{}COMP has zero complete pairs (NA). Entropy conclusions are restricted to fully aligned English cells. Complete-pair filtering requires status=ok and aligned logprobs in both arms; intervals are ordinary 95\% percentile bootstrap CIs (B=5,000), and p-values are unadjusted two-sided bootstrap p-values, distinct from the Manski-bounded, simultaneous 98.75\%, Holm-adjusted confirmatory family.

\bibliographystyle{acl_natbib}
\bibliography{references}
\end{document}